%% file: main.tex
\documentclass[runningheads]{llncs}

\usepackage{eccv}

\usepackage{eccvabbrv}

\usepackage{graphicx}
\usepackage{tikz}
\usepackage{comment}
\usepackage{amsmath,amssymb} %
\usepackage{color}
\usepackage{tabularx}
\usepackage{booktabs}
\usepackage{colortbl}
\usepackage{tabu}
\usepackage{multirow, makecell}
\usepackage{setspace}
\usepackage{pifont}
\usepackage[dvipsnames]{xcolor}

\usepackage[moderate]{savetrees}

\usepackage[font=small,labelsep=period]{caption}
\usepackage[accsupp]{axessibility}  %

\usepackage[pagebackref,breaklinks,colorlinks,citecolor=eccvblue]{hyperref}

\usepackage{orcidlink}

\definecolor{bblue}{rgb}{0.0,0.25,0.75}
\definecolor{ccol}{rgb}{0.2,0.2,0.2}

\newcommand{\myparagraph}[1]{\vspace{0.7em}\noindent\textbf{#1}}

\usepackage[capitalize]{cleveref}
\crefname{section}{Sec.}{Secs.}
\Crefname{section}{Section}{Sections}
\Crefname{table}{Table}{Tables}
\crefname{table}{Tab.}{Tabs.}

\usepackage{pifont}
\definecolor{darkergreen}{RGB}{73,157,94}

\newcommand{\cmark}{\textcolor{darkergreen}{\ding{51}}}%
\newcommand{\xmark}{\textcolor{red}{\ding{55}}}%

\DeclareMathOperator*{\argmin}{arg\,min} %

\newcolumntype{Y}{>{\centering\arraybackslash}X}
\newcolumntype{R}{>{\raggedleft\arraybackslash}X}
\newcolumntype{L}{>{\raggedright\arraybackslash}X}

\definecolor{lightgray}{rgb}{0.835, 0.835, 0.835}
\definecolor{lightergray}{rgb}{0.935, 0.935, 0.935}

\makeatletter
\def\adl@drawiv#1#2#3{%
        \hskip.5\tabcolsep
        \xleaders#3{#2.5\@tempdimb #1{1}#2.5\@tempdimb}%
                #2\z@ plus1fil minus1fil\relax
        \hskip.5\tabcolsep}
\newcommand{\cdashlinelr}[1]{%
  \noalign{\vskip\aboverulesep
           \global\let\@dashdrawstore\adl@draw
           \global\let\adl@draw\adl@drawiv}
  \cdashline{#1}
  \noalign{\global\let\adl@draw\@dashdrawstore
           \vskip\belowrulesep}}
\makeatother

\makeatletter
\newlength{\qrr@dimen@}
\expandafter\pretocmd\csname tabular*\endcsname{\setlength{\qrr@dimen@}{#1}}{}{}
\newcommand*{\Rowcolor}[2][\tabcolsep]{%
    \ifx\relax#1\relax\else
        \kern-\the\dimexpr#1\relax
    \fi
    \makebox[0pt][l]{%
        \fboxsep=0pt
        \colorbox{#2}{%
            \strut\kern\qrr@dimen@
        }%
    }%
    \ifx\relax#1\relax\else
        \kern\the\dimexpr#1\relax
    \fi
    \ignorespaces
}
\makeatother

\begin{document}

\title{\texorpdfstring{FoundPose: Unseen Object Pose Estimation\\with Foundation Features}{FoundPose: Unseen Object Pose Estimation with Foundation Features}}

\titlerunning{FoundPose: Unseen Object Pose Estimation with Foundation Features}

\newcommand{\namesep}{\hspace{0.7em}}
\author{
Evin P{\i}nar \"{O}rnek$^{1*}$ \namesep Yann Labb{\'e}$^{2}$ \namesep Bugra Tekin$^{2}$ \namesep Lingni Ma$^{2}$\\
Cem Keskin$^{2}$ \namesep Christian Forster$^{2}$ \namesep Tomas Hodan$^{2}$ 
}

\authorrunning{\"{O}rnek et al.}

\institute{
 {$^{1}$Technical University of Munich}\namesep
 {$^{2}$Meta Reality Labs}
}

\maketitle

{\let\thefootnote\relax\footnotetext{*Work done during Evin's internship at Meta.}}

\input{sections/0_abstract}

\input{sections/1_intro}

\input{sections/2_related_work}

\input{sections/3_method}

\input{sections/4_experiments}

\input{sections/5_conclusion}

\bibliographystyle{splncs04}
\bibliography{main}
\end{document}

%% file: sections/0_abstract.tex
\begin{abstract}
We propose FoundPose, a model-based method for 6D pose estimation of unseen objects from a single RGB image. The method can quickly onboard new objects using their 3D models without requiring any object- or task-specific training.
In contrast, existing methods typically pre-train on large-scale, task-specific datasets in order to generalize to new objects and to bridge the image-to-model domain gap.
We demonstrate that such generalization capabilities can be observed in a recent vision foundation model trained in a self-supervised manner.
Specifically, our method estimates the object pose from image-to-model 2D-3D correspondences, which are established by matching patch descriptors from the recent DINOv2 model between the image and pre-rendered object templates. We find that reliable correspondences can be established by kNN matching of patch descriptors from an intermediate DINOv2 layer. Such descriptors carry stronger positional information than descriptors from the last layer, and we show their importance when semantic information is ambiguous due to object symmetries or a lack of texture.
To avoid establishing correspondences against all object templates, we develop an efficient template retrieval approach that integrates the patch descriptors into the bag-of-words representation and can promptly propose a handful of similarly looking templates. Additionally, we apply featuremetric alignment to compensate for discrepancies in the 2D-3D correspondences caused by coarse patch sampling.
The resulting method noticeably outperforms existing RGB methods for refinement-free pose estimation on the standard BOP benchmark with seven diverse datasets and can be seamlessly combined with an existing render-and-compare refinement method to achieve RGB-only state-of-the-art results. Project page:
\href{https://evinpinar.github.io/foundpose/}{evinpinar.github.io/foundpose}.

\end{abstract}

%% file: sections/1_intro.tex
\section{Introduction}
\label{sec:intro}

\input{tex/teaser}

Image-based estimation of the 6D object pose (3D rotation and 3D translation) is an important research problem in the field of spatial AI.
In robotics, for example, the information about object poses allows a robot to act upon the objects, which enables fully automated solutions for warehouse operation or assembly.
In mixed-reality applications, this information unlocks physical interaction with replicas of real-world objects, such as a computer keyboard, for effective text input when fully immersed.

In this work, we address the problem of model-based 6D pose estimation of \textit{unseen} objects.
We assume that 3D models of the objects are available and that budget for onboarding the objects is limited (\eg, not sufficient for rendering a large-scale dataset and training a neural network). This is a practical problem setup for many applications
since efficient object onboarding is often a key requirement and 3D object models can be obtained from the manufacturer or readily reconstructed~\cite{newcombe2011kinectfusion,reizenstein2021common,wu2023multiview}.

The very first methods for object pose estimation can, in fact, handle unseen objects as they do not require any training, typically just a set of reference images. These methods rely on classical techniques such as matching hand-crafted image features~\cite{roberts1963machineperception,lowe1999object,ponce2004toward,collet2011moped}
or template matching ~\cite{murase1995visual,hinterstoisser2010dominant}.
Later, with the rise of machine learning techniques in computer vision, most object pose estimation methods started to rely on deep neural networks. This shift brought a significant improvement in pose estimation accuracy~\cite{sundermeyer2022bop} but limited generalization capability as large numbers of training images and a lengthy training process are usually necessary for every new object instance or category. As a result, the majority of these methods focus on a small set of objects. Only recently, with the accuracy scores of seen object pose estimation slowly saturating, the research field started to focus again on unseen objects~\cite{3dczsl2022},
with the first attempts achieving noticeably lower accuracy scores while being computationally more demanding~\cite{shugurov2022osop,labbe2022megapose,chen2023zeropose,hodan2023bop}.

With their impressive generalization capabilities, foundation models~\cite{bommasani2021opportunities} provide a solid ground for solving the problem at hand. Models such as DINOv2~\cite{oquab2023dinov2,darcet2023vision}, CLIP\cite{radford2021learning} or ALIGN~\cite{jia2021scaling} have been successfully applied on various vision tasks without any task-specific training~\cite{melaskyriazi2022deep,wang2022selfsupervised,goodwin2022,nguyen2023cnos}. For example, CNOS~\cite{nguyen2023cnos} leverages frozen DINOv2~\cite{oquab2023dinov2,darcet2023vision} and Segment Anything~\cite{kirillov2023segany} and outperforms Mask R-CNN~\cite{he2017mask} on the object segmentation task. Goodwin \etal\cite{goodwin2022} show that DINO patch descriptors\cite{caron2021emerging} can be used to establish semantic correspondences between instances of the same object category.

Inspired by these success stories, we propose FoundPose, a method for model-based pose estimation of unseen objects,
which brings the power of modern foundation features into classical computer vision techniques via careful design choices. Despite being surprisingly simple, easy-to-interpret, and requiring no object- or task-specific training, the method achieves state-of-the-art results on the standard BOP benchmark~\cite{hodan2023bop}.

First, given an RGB image and an object mask from CNOS~\cite{nguyen2023cnos}, we perspectively crop the object region and rapidly retrieve a small set of similarly looking, pre-rendered object templates.
To this end, we develop an efficient retrieval approach by integrating DINOv2 into the \emph{bag-of-words} representation from 2003~\cite{sivic2003video}. This approach is 15X faster while only slightly less accurate than the heavy render-and-compare coarse stage of MegaPose~\cite{labbe2022megapose}, and requires 100X less templates (several hundreds vs 90K+) than previous approaches~\cite{shugurov2022osop,nguyen2022template} (the overall memory footprint is 25X lower).

Second, we establish 2D-2D correspondences between each retrieved \emph{synthetic template} (with fixed light and black background) and the \emph{real image crop} by simple one-way kNN search of DINOv2 patch descriptors (Fig.~\ref{fig:teaser}). In contrast, existing methods typically train on large-scale, heavily randomized, and task-specific datasets to bridge the synthetic-to-real domain gap~\cite{hodan2020bop,labbe2022megapose,sundermeyer2018implicit}. We demonstrate that patch descriptors from an \emph{intermediate} DINOv2 layer, which were shown to carry stronger positional information~\cite{amir2021deep}, are crucial for achieving geometrically consistent correspondences when semantic information is ambiguous due to object symmetries or a lack of texture. We show that the intermediate DINOv2 descriptors are in fact the key enabler of FoundPose, yielding significantly higher accuracy also compared to descriptors extracted with SAM\cite{kirillov2023segany}, CLIP~\cite{radford2021learning}, LoFTR~\cite{sun2021loftr}, S2DNet~\cite{germain2020s2dnet}, and dense SIFT~\cite{lowe1999object}.
Next, for each retrieved template, we generate a pose hypothesis from image-to-model 2D-3D correspondences, which are established by lifting the matched 2D patch locations in the template to 3D using rendered depth.
Finally, we further optimize the top-quality hypothesis by \emph{featuremetric refinement}, which applies the idea of the classical photometric refinement \cite{baker2004lucas} to DINOv2 patch descriptors. The refinement effectively compensates for the discrepancy in the 2D-3D correspondences caused by coarse sampling of DINOv2 patches.

\vspace{3ex}
\noindent
In summary, we make the following contributions:
\begin{enumerate}

\item \emph{A training-free method for model-based object pose estimation}
which relies on a surprising simple and easy-to-interpret DINOv2-based pipeline and achieves state-of-the-art results on the standard BOP benchmark~\cite{hodan2023bop}.
 
\item \emph{An efficient template retrieval approach} which requires 100X fewer templates than previous approaches and is robust to partial object occlusions.

\item \emph{A lightweight object representation} which is fast to build and has a 25X lower memory footprint than competitors, enabling scaling to large numbers of objects.

\item \emph{A featuremetric refinement approach} which compensates for coarse patch sampling.

\item \emph{Demonstrated importance of intermediate DINOv2 descriptors} for handling symmetric and texture-less objects, also outperforming descriptors from other foundation models.

\end{enumerate}

%% file: tex/teaser.tex
\begin{figure}[t]
  \begin{minipage}[t][][c]{0.5\textwidth}
    \includegraphics[width=\textwidth]{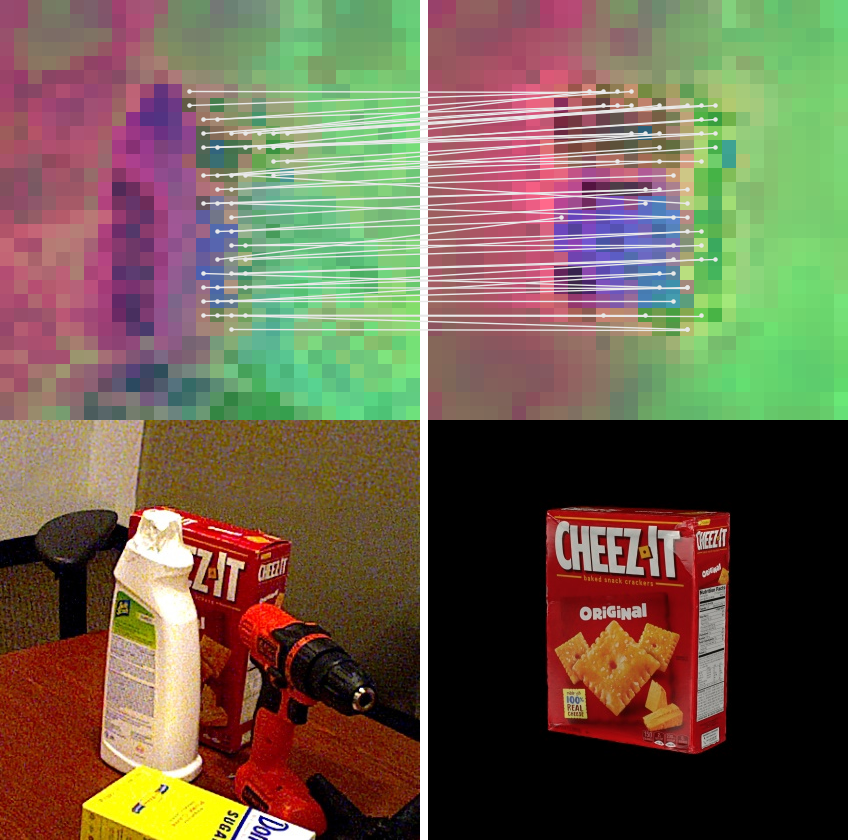}
  \end{minipage}
  \hfill
  \begin{minipage}[t][][c]{0.47\textwidth}
  \vspace{0.93em}
  \caption{
   \textbf{Bridging synthetic-to-real~gap.}
   Patch descriptors from
   an intermediate layer of
   DINOv2\cite{oquab2023dinov2} (top), a recent vision foundation model, are the key enabler of FoundPose.
   Thanks to the generalization capability of these descriptors, it is possible to establish reliable correspondences between a \emph{real query image} (left) and a \emph{synthetic template} (right) by a simple nearest-neighbor matching.
   The patch descriptors are colored by the top three components of a PCA space computed from descriptors of all object templates. Note that colors of the same object parts are consistent, despite the real-to-synthetic domain gap.
   }
   \label{fig:teaser}
  \end{minipage}
\end{figure}

%% file: sections/2_related_work.tex
\section{Related work}
\label{sec:related}
This paper builds on over 60 years of research in object pose estimation and on the recent large-scale vision foundation models.

\myparagraph{Classical methods.} Estimating the 6D pose of rigid objects from a single image is one of the first computer vision problems~\cite{roberts1963machineperception}. Early methods relied on local feature matching~\cite{lowe1999object,ponce2004toward,collet2011moped} or template matching \cite{murase1995visual,hinterstoisser2010dominant}, and could rapidly onboard new objects if provided with a set of reference images annotated with model-to-camera transformations. With the introduction of Microsoft Kinect, the attention of the research field was steered towards object pose estimation from RGB-D or D-only images, yielding methods based on 3D local features\cite{tombari2013performance}, notably successful point-pair features\cite{drost2010model,hodan2018bop}, and methods based on RGB-D template matching\cite{hinterstoisser2012accv,hodan2015detection}. The RGB-D methods produce more accurate poses and are therefore popular in industry, but their application in open-world scenarios is limited. Besides boosting the pose estimation accuracy, the additional depth channel from Kinect-like sensors enabled easy 3D object reconstruction~\cite{newcombe2011kinectfusion}, and, in turn, methods relying on 3D mesh models started to emerge. The model-based object pose estimation setup is still popular~\cite{sundermeyer2022bop,hodan2023bop}, among both RGB and RGB-D methods, and is relevant for factory and warehouse scenarios where CAD object models are often available. On the other hand, the model-free setup, recently revisited in \cite{sun2022onepose,he2022oneposeplusplus}, is relevant for mixed reality applications where the set of target objects is typically small and capturing reference images is easy.

\myparagraph{Deep learning methods.}
As in other fields of computer vision, methods based on hand-crafted features and techniques have been progressively replaced
by methods based on deep neural networks\cite{xiang2018posecnn,manhardt2018deep,li2020deepim,labbe2020cosypose,wang2021gdrnet}, which can operate on RGB or RGB-D inputs. These methods represent the current state of the art in terms of accuracy~\cite{sundermeyer2022bop,hodan2023bop}. However, their scalability is hindered by the requirement of a large-scale training dataset for learning new objects.
To address this issue, deep-learning methods that can onboard new objects without any object-specific training have been proposed recently. As examples of model-based methods, Nguyen~\etal\cite{nguyen2022template}, Shugurov~\etal (OSOP)\cite{shugurov2022osop}, and Thalhammer~\etal\cite{thalhammer2023selfsupervised} learn descriptors for template matching by contrastive learning, Sundermeyer~\etal\cite{sundermeyer2020multipath} generate such descriptors by an augmented auto-encoder,
Pitteri~\etal \cite{pitteri2019cornet,pitteri20203d} predict generic 3D keypoints or local surface embeddings, and Xiao~\etal\cite{Xiao2019PoseFromShape} directly predict the 3D object orientation.
Model-free methods~\cite{he2022fs6d,liu2022gen6d,sun2022onepose,he2022oneposeplusplus}, and methods relying on depth measurements~\cite{park2020latentfusion,Balntas2017PoseGR,okorn2021zephyr,chen2023zeropose} have been also proposed.

The top-performing methods on the unseen object pose estimation task of the BOP Challenge 2023~\cite{hodan2023bop} include GenFlow\cite{genflow}, MegaPose\cite{labbe2022megapose}, GigaPose\cite{nguyen2024gigaPose}, FoundationPose\cite{wen2023foundationpose}, SAM-6D\cite{lin2023sam}, and PoMZ\cite{caraffa2023object}. To achieve generalization to novel objects, all except PoMZ require generating millions of task-specific training images showing thousands of different objects. Generating such datasets requires significant effort and opens up new types of challenges, including positioning objects in the scene\cite{hodan2019photorealistic,denninger2020blenderproc}, collecting a sufficiently large set of object models, or texturing the models\cite{wen2023foundationpose}. In contrast, FoundPose does not require any training, uses frozen DINOv2 features, and outperforms RGB-only GenFlow, MegaPose, and GigaPose. FoundationPose, SAM-6D, PoMZ, as well as ZeroPose\cite{chen2023zeropose}, are RGB-D methods and do not have RGB-only versions.
To the best of our knowledge, the only training-free methods for model-based object pose estimation are PoMZ\cite{caraffa2023object}, which requires RGB-D inputs, and ZS6D\cite{Ausserlechner2023ZS6DZ6}, which achieves significantly lower accuracy than FoundPose.
ZS6D
uses features from the last ViT~\cite{dosovitskiy2021image}
layer for establishing correspondences, which we show inferior to our solution (Sec.~\ref{sec:exp}).
Note that \cite{genflow,chen2023zeropose,Ausserlechner2023ZS6DZ6,caraffa2023object,nguyen2024gigaPose,wen2023foundationpose} are all unpublished at the time of submission.

\myparagraph{Foundation models.}
A foundation model is a machine-learning model trained on broad data by self-supervised learning that can be adapted to a wide range of downstream tasks~\cite{bommasani2021opportunities}. Foundation models initially appeared in natural language processing with examples such as BERT~\cite{devlin2019bert} or GPT-3\cite{brown2020language}. In computer vision, foundation models already achieve on-par or better results than supervised models
\cite{bommasani2021opportunities,caron2021emerging,oquab2023dinov2,cherti2023reproducible,wang2022selfsupervised,melaskyriazi2022deep,ornek23}. A prominent example is DINOv2~\cite{oquab2023dinov2}, which is based on the Vision Transformer architecture\cite{dosovitskiy2021image}, trained in a self-distillation fashion, and has been shown to encode fine spatial information about the object parts as well as semantic information about object categories~\cite{amir2021deep}. It has been successfully used in zero-shot setups, \ie, without any training, for establishing semantic correspondences \cite{amir2021deep,zhang2023tale,goodwin2022,liu2023matcher}.
FoundPose builds on these insights.
Compared to Goodwin \etal \cite{goodwin2022} which use DINO patch descriptors to establish semantic correspondences within an object category, FoundPose shows that DINOv2 descriptors can be used to establish synthetic-to-real correspondences. Furthermore, \cite{goodwin2022} requires RGB-D inputs at test time, cannot handle symmetric objects (such objects are omitted in their evaluation), and is compared only with custom baselines. FoundPose assumes RGB-only test inputs, can handle symmetric objects by design, and achieves state-of-the-art results on the standard BOP benchmark. From the already reviewed methods for pose estimation of specific objects, PoMZ\cite{caraffa2023object} and ZS6D\cite{Ausserlechner2023ZS6DZ6} also rely on frozen DINOv1/v2. However, the first needs RGB-D inputs, and the latter achieves noticeably lower accuracy.

%% file: sections/3_method.tex
\section{FoundPose}

\label{sec:method}

In this section, we
describe FoundPose, the proposed method for unseen object pose estimation. We first provide a high-level overview of the method
in Sec.~\ref{sec:method_overview} and then focus on the key components and rationale of our design choices in Sec.~\ref{sec:object_repre}--\ref{sec:feature_align}.

\subsection{Method overview} \label{sec:method_overview}

\noindent\textbf{Problem definition.}
We consider the problem of estimating the 6D pose of rigid objects from a single RGB image with known intrinsics. The objective is to estimate the pose of all instances of target objects that are visible in the image. We assume that the only information provided for the target objects are their 3D mesh models and that there is only a limited budget for onboarding the objects,
\ie, for preparing object representations that can be used for online pose estimation. We constrain the onboarding process to 5 minutes and 1 GPU, as required by the BOP Challenge 2023\cite{hodan2023bop}.
We additionally assume that segmentation masks of the target object instances, together with per-mask object identity, are provided at inference time. In our experiments, we obtain the masks by CNOS~\cite{nguyen2023cnos}, a recent method for segmentation of unseen objects that
also requires only 3D models for onboarding the objects.

\myparagraph{Onboarding and inference.}
During an offline onboarding stage, we render templates showing 3D object models in different orientations. From each template, we extract DINOv2 descriptors of image patches
and register the descriptors in 3D, \ie, each patch descriptor is associated with the corresponding 3D location in the object model space (Sec.~\ref{sec:object_repre}).
At inference time, given a segmentation mask of an object instance, we crop the image region around the mask, extract DINOv2 patch descriptors of the crop, and apply a bag-of-words retrieval technique to efficiently identify a small set of templates that show the object in orientations similar to the observation (Sec.~\ref{sec:template_retrieval}). For each retrieved template, we establish 2D-3D correspondences by matching patch descriptors from the crop against patch descriptors from the template, 
and generate a pose hypothesis by the P\emph{n}P-RANSAC algorithm (Sec.~\ref{sec:2d3d_corresp}). Finally, we refine the best pose hypothesis by featuremetric alignment, an optimization-based algorithm inspired by photometric alignment that operates on features (Sec.~\ref{sec:feature_align}). The pipeline of the method is shown in Fig.~\ref{fig:pipeline}.

\input{tex/overview}

\subsection{Template-based object representation} \label{sec:object_repre}

\noindent\textbf{Template generation.} Given a texture-mapped 3D object model, we render $n$ RGB-D templates showing the model under different orientations. The orientations are sampled to uniformly cover the $SO(3)$ group of 3D rotations~\cite{alexa2022super}, and the model is rendered using a standard rasterization technique~\cite{shreiner2009opengl} with a black background and fixed lighting. The size of templates is $S\times S$ pixels, and the objects are rendered such that the longer side of their 2D bounding box is $\delta S$ pixels long, with $\delta < 1$. At inference, we generate crops of the query image with the same size
and padding (to allow for errors of segmentation masks around which we crop the image). %

\myparagraph{Patch descriptors registered in 3D.}
For each RGB-D template with an index $t \in \{1, \dots , n\}$, we split the RGB channels into $m$ non-overlapping patches of $14\times14$ pixels and calculate their patch descriptors $\{\mathbf{p}_{t,i}\}_{i=1}^m$. A patch descriptor is calculated as $\mathbf{p}_{t,i} = \phi_d(\mathbf{p}'_{t,i})$, where $\mathbf{p}'_{t,i}$ is the raw patch descriptor extracted by DINOv2 and $\phi_d: \mathbb{R}^r \mapsto \mathbb{R}^d$ projects the $r$-dimensional raw descriptor to the top $d$ PCA components, which are calculated from valid patch descriptors of all $n$ templates. A patch is considered valid if its 2D center falls inside the object mask and the PCA-based dimensionality reduction is applied to increase efficiency.
Then we represent a template $t$ by a set $T_t = \{(\mathbf{p}_{t,j}, \mathbf{x}_j) \; | \; j \in M \}$, where $M$ are indices of valid patches, $\mathbf{x}_j$ is a 3D location (in the coordinate space of the 3D object model) whose 2D projection is at the center of patch~$j$. The 3D locations, which are calculated from the depth channel of the template and known camera intrinsics, enable establishing 2D-3D correspondences at inference.

\myparagraph{Bag-of-words descriptors.} 
At onboarding, we also pre-calculate bag-of-words descriptors of all templates to enable efficient template retrieval at inference using the classical bag-of-words image retrieval technique~\cite{sivic2003video,philbin2008lost}, which mimics text-retrieval systems with the analogy of visual words.
Specifically, we define visual words as the centroids of $k$-means clusters of patch descriptors extracted from all templates of an object. To calculate the bag-of-words descriptor of a template $t$, we assign patch descriptors from the template representation $T_t$ to the nearest visual words and describe the template by a vector $\mathbf{b}_t = (b_1, b_2, \dots, b_k)$. This vector consists of weighted word frequencies defined as $b_i = (n_{i,t}/n_t) \log(N/n_i)$, where $n_{i,t}$ is the number of occurrences of word $i$ in template $t$, $n_t$ is the total number of words in template $t$, and $n_i$ is the number of occurrences of word $i$ in all $N$ templates. The first term ($n_{i,t}/n_t$) weights words that occur often in a particular template and therefore describe the template well, while the second term ($\log(N/n_i)$) downweights words that occur often in any template.
As visual words generated by clustering may suffer from quantization errors, we follow \cite{philbin2008lost,torii2013visual} and soft-assign each patch descriptor to several nearest words with weights defined by $\exp(-d^2/2\sigma^2)$, where $d$ is the Euclidean distance of the descriptor from the word and $\sigma$ is a parameter
of the method.

\subsection{Template retrieval by bag-of-words matching} \label{sec:template_retrieval}

\noindent\textbf{Perspective cropping.}
At inference, we start by cropping the image region around a given object segmentation mask. To minimize perspective distortion and achieve a crop that resembles a template, we generate the crop by warping the query image to a virtual pinhole camera focused on the segmentation mask. The virtual camera is constructed such that its optical axis passes through the center of the 2D bounding box of the mask, the viewport size is $S \times S$ pixels, and the longer side of the warped 2D bounding box is $\delta S$ pixels long.

\myparagraph{Retrieving similar templates.}
To retrieve a small set of templates, we calculate the bag-of-words descriptor of the crop (as in Sec.~\ref{sec:object_repre}) and calculate its cosine similarity (\ie, normalized scalar product) with bag-of-words descriptors of all object templates. We select $h$ templates with the highest cosine similarity, which provide approximate hypotheses on the object orientation for the subsequent pose estimation stage. 

This retrieval technique is efficient and robust to partial occlusions.
When an object is partially occluded, its visible part still contributes visual words describing the object.
The cosine similarity then normalizes the magnitude and focuses on the direction of the bag-of-words descriptors and is, therefore, less sensitive to the number and more to the type of present words. This robustness has been described in prior work \cite{philbin2007object,sivic2003video} and also in our experiments (see results on LM-O and T-LESS in Sec.~\ref{sec:exp}).

Since the bag-of-words descriptor represents an image as a bag of unordered visual words, the retrieval can be typically improved by re-ranking the results with a spatial verification stage~\cite{philbin2007object}. However, in our case, a similar verification is implicitly done by the subsequent P\emph{n}P-RANSAC (spatially consistent correspondences are expected to yield a better pose estimate), and bags of words are not actually unordered as the used patch descriptors from an intermediate DINOv2 layer, from which the words are constructed, contains 2D positional information.

\input{tex/dino_feats}

\subsection{Pose estimation from 2D-3D correspondences} \label{sec:2d3d_corresp}

\noindent\textbf{Crop-to-template patch matching.}
For each retrieved template~$t$, we match patch descriptors
from the crop to the nearest descriptors from $T_t$ (in terms of the Euclidean distance),
and establish 2D-3D correspondences $C_t = \left\{ \left( \mathbf{u}_i, \mathbf{x}_i\right) \right\}_{i=1}^m$, where $\mathbf{u}_i$ is the 2D center of a query patch and $\mathbf{x}_i$ is the 3D location associated with the matched patch descriptor from~$T_t$. The cyclic matching from Goodwin \etal \cite{goodwin2022} did not help in our setup.

Establishing 2D-3D correspondences by crop-to-template patch matching
is a considerably simpler problem than exhaustive matching against patch descriptors from all templates, which would be necessary without the template retrieval stage. Moreover, we demonstrate that the template-based approach can effectively handle arbitrary objects, including challenging objects with symmetries and without a significant texture. The ambiguity of 2D-3D correspondences, for which such objects are notoriously known~\cite{hodan2020epos}, is eliminated by (1) restricting the candidate patches to only a single template and (2)~using patch descriptors from an intermediate layer of DINOv2 which contain both semantic and 2D positional information~\cite{amir2021deep}. We find that the positional information is crucial for producing geometrically consistent correspondences when the semantic information is not discriminative due to symmetries or a lack of texture (see Fig.~\ref{fig:feature_vis} and Sec.~\ref{sec:ablations}).

\myparagraph{Pose fitting.} An object pose $(\mathbf{R}_t, \mathbf{t}_t)$, defined by a 3D rotation $\mathbf{R}_t$ and a 3D translation $\mathbf{t}_t$ from the model space to the camera space, is estimated for each retrieved template $t$ from 2D-3D correspondences $C_t$ by solving the Perspective-\emph{n}-Point (P\emph{n}P) problem. We solve this problem by the EP\emph{n}P algorithm~\cite{lepetit2009epnp} combined with the RANSAC fitting scheme~\cite{fischler1981random} for robustness. In this scheme, P\emph{n}P is solved repeatedly on a randomly sampled minimal set of 4 correspondences, and the final output is defined by the pose hypothesis with the highest quality, which we define by the number of inlier correspondences~\cite{fischler1981random}. From the set of $h$ poses estimated from the $h$ retrieved templates, the pose with the highest quality is selected as the final coarse pose estimate $(\mathbf{R}_c, \mathbf{t}_c)$.

\subsection{Featuremetric pose refinement} \label{sec:feature_align}

\noindent\textbf{Reducing 2D-3D discrepancy.} The 2D-3D correspondences are established by
linking the 2D centers of matched crop and template patches and lifting the centers of the template patches to 3D using the rendered depth channel. Since the patches are relatively large ($14 \times 14$ px), the 2D centers of the crop patches may not precisely align with 2D projections of the corresponding 3D points, even when projected using the ground-truth model-to-crop transformation. To compensate for the potential discrepancy, we refine the pose estimates by aligning template patches to their optimal locations in the image crop by a featuremetric optimization described below.

\myparagraph{Featuremetric alignment.} The coarse pose $(\mathbf{R}_c, \mathbf{t}_c)$ is refined by Levenberg-Marquardt (L-M) \cite{levenberg1944,marquardt1963}, an iterative non-linear optimization algorithm.
Each L-M iteration updates the pose parameters by a gradient-descent step, minimizing a real-valued cost while adaptively selecting between the first and second-order gradients depending on the cost value. We initialize the optimization with the coarse pose $(\mathbf{R}_c, \mathbf{t}_c)$ and obtain a refined pose $(\mathbf{R}_r, \mathbf{t}_r)$ by minimizing the following featuremetric error:

\begin{align*}
(\mathbf{R}_r, \mathbf{t}_r) = \argmin_{(\mathbf{R}, \mathbf{t})} \sum_{(\mathbf{p}_{i}, \mathbf{x}_{i}) \in T_t} \rho \Big( \mathbf{p}_{i} - \textbf{F}_q\Big(\pi(\mathbf{R}\mathbf{x}_{i}+\mathbf{t}) / s \Big) \Big),
\end{align*}
\vspace{0.5ex}

\noindent where $\rho$ is a robust cost function by Barron~\cite{barronloss}, $(\mathbf{p}_{i}, \mathbf{x}_{i}) \in T_t$ is a descriptor and the corresponding 3D location of patch $i$ from template $t$, and $\pi: \mathbb{R}^3 \mapsto \mathbb{R}^2$ represents the 2D projection. The feature map $\textbf{F}_q \in \mathbb{R}^{a \times a \times d}$ is obtained by stacking the patch descriptors of the query image and is sampled with bilinear interpolation at normalized 2D projections. 
The spatial resolution of the feature map is $a \times a$ with $a=S/s$, where $s \times s$ is the patch size.
Note that a similar featuremetric alignment was applied in several works \cite{stumberg2020, sarlin21pixloc}, typically in combination with features trained specifically for L-M. In our work, we apply the alignment directly on DINOv2 features without any training. Even though there is no guarantee for reaching the global optimum, as in other methods optimizing photo/featuremetric objectives, we observed that a plausible solution is often reached.

%% file: tex/overview.tex
\begin{figure*}[t]
  \centering
    \includegraphics[width=1.0\linewidth]{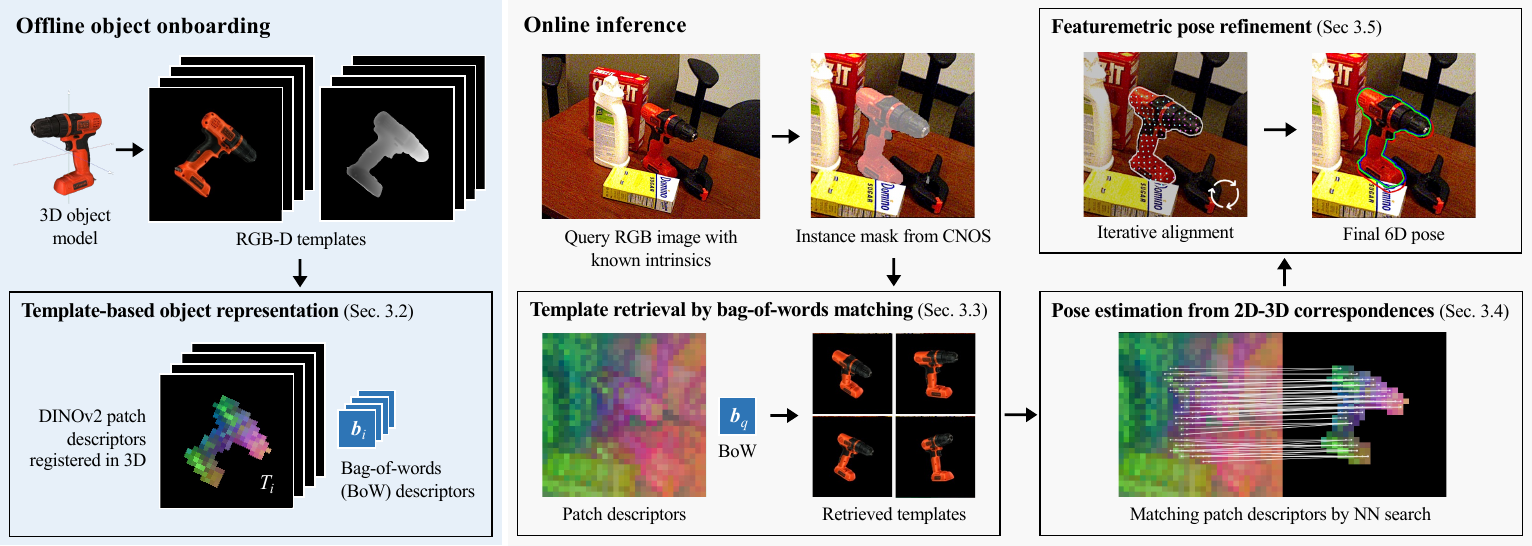}
   \caption{\textbf{FoundPose overview.} During a short onboarding stage, we render RGB-D templates showing the object in different orientations, extract DINOv2 patch descriptors~\cite{oquab2023dinov2,darcet2023vision} from the RGB channels and register the descriptors in 3D using the depth channel. At inference time, we crop the RGB query image around the object mask predicted by CNOS~\cite{nguyen2023cnos} and retrieve a small set of most similar templates using a bag-of-words approach (with words defined by \emph{k}-means clusters of patch descriptors from all templates). For each retrieved template, a pose hypothesis is generated by P\emph{n}P-RANSAC~\cite{fischler1981random,lepetit2009epnp} from 2D-3D correspondences established by matching patch descriptors of the image crop and the template. Finally, the pose hypothesis with the highest number of inlier correspondences is refined by featuremetric alignment.
   }
   \label{fig:pipeline}
   \vspace{-0.5em}
\end{figure*}

%% file: tex/dino_feats.tex
\begin{figure}[t!]
  \begin{center}
  \includegraphics[width=0.7\linewidth] {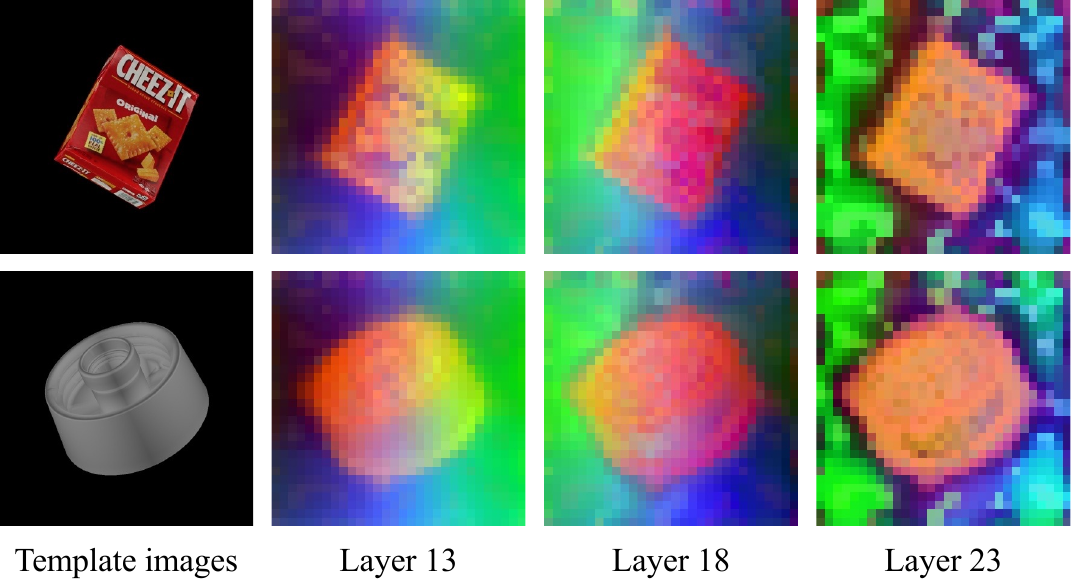}
  \caption{\textbf{Visualization of DINOv2 patch descriptors.}  Shown are top three PCA components of patch descriptors from different layers of DINOv2 ViT-L\cite{darcet2023vision}, for a textured object from YCB-V\cite{xiang2018posecnn} (top) and a symmetric and texture-less object from T-LESS~\cite{hodan2017tless} (bottom). As observed in\cite{amir2021deep} and also clearly visible in these visuals, the patch descriptors contain gradually less positional and more semantic information when going from shallower to deeper layers -- the different coloring of object sides (red left vs yellow right) in Layer 13 gradually blends to a solid color (orange) in Layer 23. %
  FoundPose performs the best with descriptors from layer 18, which presumably provides the right information mix. We observed that these descriptors produce geometrically consistent correspondences even on symmetric and texture-less objects -- when the semantic information is ambiguous (due to symmetries or a lack of texture), the positional information prioritizes matching patches from the same object side.
   }\label{fig:feature_vis}
   \vspace{-1.0em}
   \end{center}
\end{figure}

%% file: sections/4_experiments.tex
\section{Experiments}
\label{sec:exp}

In this section, we compare the accuracy and speed of FoundPose with the state-of-the-art methods evaluated on the BOP benchmark~\cite{hodan2018bop,hodan2020bop,sundermeyer2022bop,hodan2023bop} and present ablation experiments.

\input{tex/quals}

\subsection{Experimental setup} \label{sec:exp_setup}

\noindent \textbf{Evaluation protocol.} We follow the protocol of the BOP Challenge 2019--2023~\cite{hodan2020bop}. In summary, a method is evaluated on the 6D object localization problem, and the error of an estimated pose \wrt the ground-truth pose is calculated by three pose-error functions:
Visible Surface Discrepancy (VSD) treats ambiguous poses as equivalent by considering only the visible object part, Maximum Symmetry-Aware Surface Distance (MSSD) considers a set of pre-identified global object symmetries and measures the surface deviation in 3D, and Maximum Symmetry-Aware Projection Distance (MSPD) considers the object symmetries and measures the perceivable deviation.
An estimated pose is considered correct \wrt a pose-error function~$e$, if $e < \theta_e$, where $e \in \{\text{VSD}, \text{MSSD}, \text{MSPD}\}$ and $\theta_e$ is the threshold of correctness. The fraction of annotated object instances for which a correct pose is estimated is referred to as Recall. The Average Recall \wrt a function~$e$, denoted as $\text{AR}_e$, is defined as the average of the Recall rates calculated for multiple settings of the threshold $\theta_e$ and also for multiple settings of a misalignment tolerance $\tau$ in the case of $\text{VSD}$. The overall accuracy of a method is measured by the Average Recall:
$\text{AR} = (\text{AR}_{\text{VSD}} + \text{AR}_{\text{MSSD}} + \text{AR}_{\text{MSPD}}) \, / \, 3$.

\myparagraph{Datasets.} The experiments are conducted on the seven core BOP datasets: LM-O~\cite{brachmann2014learning}, T-LESS~\cite{hodan2017tless}, ITODD~\cite{drost2017introducing}, HB~\cite{kaskman2019homebreweddb}, YCB-V~\cite{xiang2018posecnn}, IC-BIN~\cite{doumanoglou2016recovering}, and TUD-L~\cite{hodan2018bop}. The datasets feature 108 diverse objects ranging from texture-less and symmetric industrial objects (ITODD, T-LESS) to typical household objects. The images show scenes whose complexity varies from simple scenes with several isolated objects to challenging ones with multiple object instances and a high amount of clutter and occlusion. Only 3D object models and test images from these datasets were used for experiments with FoundPose, not the provided synthetic nor real training images since no training is required.

\myparagraph{Compared methods.} FoundPose is compared against model-based RGB methods evaluated on the unseen object pose estimation task of the BOP Challenge 2023~\cite{hodan2023bop}: GenFlow\cite{genflow}, MegaPose~\cite{labbe2022megapose}, GigaPose \cite{nguyen2024gigaPose}, and also against ZS6D~\cite{Ausserlechner2023ZS6DZ6} and OSOP~\cite{shugurov2022osop}.
Except for OSOP, all of these methods (including FoundPose) use the same segmentation masks that were produced by CNOS~\cite{nguyen2023cnos} and provided to the challenge participants.
OSOP relies on a custom detector of unseen objects for LM-O, HB, and YCB-V, and on Mask R-CNN~\cite{he2017mask} trained for specific objects for T-LESS (hence we do not include the T-LESS result).
Besides variants of FoundPose where the coarse poses are refined by the featuremetric refinement (Sec.~\ref{sec:feature_align}), we evaluate variants with poses refined by 5 iterations of the MegaPose refiner (\ie, the last stage of the MegaPose pipeline~\cite{labbe2022megapose}).

\myparagraph{Implementation details.} Unless stated otherwise, we use the following parameter settings in the presented experiments. We rendered 800 templates per object with approximately $25^{\circ}$ angle between depicted object orientations. We set the size of templates and of the query image crop to $420 \times 420$ px with $\delta=0.6$. With the patch size of $14 \times 14$ px (for which DINOv2 is trained), we extract $30 \times 30$ patch descriptors from each template/crop and reduce their dimensionality by projecting them to the top 256 PCA components. We use the output tokens from layer 18 of DINOv2 ViT-L/14 with registers\cite{darcet2023vision} as the patch descriptors. Visual words for the bag-of-words template retrieval are defined per object by the centroids of 2048 $k$-means clusters of patch descriptors from all templates of the object. The bag-of-words descriptors are constructed by soft-assigning each patch descriptor to 3 nearest words with $\sigma=10$. For each query image crop, we retrieve 5 templates,
and estimate the pose from 2D-3D correspondences (established between the query image crop and the template) by P\emph{n}P-RANSAC running for up to 400 iterations with the inlier threshold set to $10$ px. The featuremetric refinement is applied to the best coarse pose and runs until convergence for up to 30 iterations, with the Barron loss~\cite{barronloss} parameters set to $\alpha=-5$ and $c=0.5$.
By default, the evaluated FoundPose variants use $n$ CNOS masks per object, where $n$ is the number of object instances to localize (provided as input in the 6D object localization task in BOP). The only exceptions are variants in the bottom part of Tab.~\ref{tab:main_results}, which use 5$n$ CNOS masks per object.
Note that all masks were loaded from files with default CNOS masks, which were provided for BOP 2023\cite{hodan2023bop}
and contain multiple masks per object instance. The number of CNOS masks used by other methods is unknown.

\subsection{Main results}
\noindent\textbf{Accuracy.} Among methods that do not apply any refinement stage, FoundPose (without the featuremetric refinement) produces significantly more accurate poses than the competitors, achieving +10, +14, and +16 AR on the seven BOP datasets on average compared to the coarse versions of GigaPose \cite{nguyen2024gigaPose}, GenFlow~\cite{genflow}, and MegaPose~\cite{labbe2022megapose} (rows 1--4 in Tab.~\ref{tab:main_results}).
The featuremetric refinement brings an extra improvement of +5 AR on average (rows 1 vs 7). At an additional computational cost, a large improvement of +17 AR (rows 1 vs 8) can be achieved if the coarse poses from FoundPose are refined by the iterative render-and-compare approach from MegaPose, which is trained on 2M+ synthetic images of diverse objects and proven remarkably effective. When initiated with coarse poses from FoundPose, the MegaPose refiner achieves +4 higher AR score than when initiated with poses from the original coarse pose estimation stage of MegaPose (rows 8 vs 11). Further improvements at further computational cost can be achieved if the refinement is applied to multiple pose hypotheses and the top refined pose is reported as the final estimate (rows 12--17).
We achieve the overall best average AR score of 59.6 AR when top 5 pose hypotheses (generated from 5 retrieved templates) are optimized with the featuremetric refinement followed by the MegaPose refinement. On both the single and the multi hypotheses setups, combining the two refinement approaches achieves the best scores (rows 9 and 14), suggesting their complementarity.
This entry outperforms multi-hypotheses versions of MegaPose and GenFlow (rows 16 and 17), which are the top-performing RGB methods from the BOP Challenge 2023~\cite{hodan2023bop}, as well as GigaPose\cite{nguyen2024gigaPose} with the MegaPose refinement (row 15).

\input{tex/mainresults}

\input{tex/ablations}

\myparagraph{Speed.} The presented variants of FoundPose are the most accurate and the second-fastest entries in all three categories evaluated in Tab.~\ref{tab:main_results}.\footnote{The time of GigaPose~\cite{nguyen2024gigaPose}, the only faster method, was obtained with a more powerful GPU (V100 48GB vs P100 16GB used in our experiments).}\footnote{OSOP~\cite{shugurov2022osop} has a light version that runs at < 1\,s per image, but has lower accuracy scores (-3 AR) than the entry in Tab.~\ref{tab:main_results}, which takes 5--12\,s per image.} Notably, the coarse pose estimation stage from FoundPose is significantly faster (1.7 vs. 15.5\,s) and more accurate than the coarse render-and-compare stage from MegaPose (rows 1 vs. 4). The presented FoundPose variants provide a spectrum of trade-offs between speed and accuracy, with the faster variants (rows 1 and 7) being relevant for online applications such as robotic manipulation, while the more accurate variants being relevant for offline applications such as ground-truth object pose annotation for training supervised methods~\cite{sundermeyer2022bop}. The trade-off can be further controlled by, \eg, the template size
or different types of the DINOv2 backbone -- rows 3--4 in Tab.~\ref{tab:ablations} show that FoundPose runs at around 1.3\,s per image if the ViT-L backbone (used for results in Tab.~\ref{tab:main_results}) is replaced with the smaller ViT-S. Note that the reported times also include the CNOS segmentation stage that takes around 0.3\,s per image. The offline object onboarding takes less than 5 minutes, as requested by BOP~\cite{hodan2023bop}.

\subsection{Ablation experiments} \label{sec:ablations}

\noindent\textbf{Feature extractors.} Tab.~\ref{tab:ablations} (rows 1--7) evaluates the performance of FoundPose (without the featuremetric refinement) with different patch descriptors. We achieve the best pose accuracy when we use the output tokens of layer 18 (out of 23) from the DINOv2 ViT-L model with registers\cite{darcet2023vision} as the patch descriptors. We also observed qualitatively that this layer provides a good balance between the positional and semantic information -- the resulting correspondences are more geometrically consistent than correspondences obtained using later layers (Fig.~\ref{fig:feature_vis}).
Furthermore, we evaluate FoundPose with patch descriptors extracted from layer 23 of SAM ViT-L~\cite{kirillov2023segany}, from the last backbone layers of the feature matching pipelines LoFTR~\cite{sun2021loftr} and S2DNet~\cite{germain2020s2dnet}, from the last layer of the CLIP image encoder~\cite{radford2021learning}, and with patch descriptors defined by the classical SIFT descriptor~\cite{lowe2004distinctive} calculated on a regular 2D grid with the cell size of 7\,px. In the case of S2DNet, descriptors were obtained by sampling the last CNN feature map. All method parameters were fixed in these experiments, only the patch descriptors differed. As shown in Tab.~\ref{tab:ablations}, patch descriptors from DINOv2 are the key enabler of FoundPose, yielding significantly higher accuracy than the alternatives (we only show the performance of the better alternatives).

\myparagraph{Template retrieval.} Next, on rows 8--9 in Tab.~\ref{tab:ablations}, we evaluate the proposed bag-of-words template retrieval approach against an alternative based on matching the\;\texttt{cls}\;token from DINOv2, which is used in CNOS~\cite{nguyen2023cnos} for object identification. In this alternative, we use the\;\texttt{cls}\;token from layer 18 of DINOv2 ViT-L as the template and crop descriptor and retrieve templates whose descriptors have the highest cosine similarity with the crop descriptor.
Compared to the bag-of-words descriptor constructed from patch descriptors, we observed that the\;\texttt{cls}\;token is less robust to occlusions and contains limited information about the object pose (as also shown in Fig.~5 of~\cite{nguyen2023cnos}). To avoid the influence of the background on the\;\texttt{cls}\;token, we additionally evaluate a variant where pixels outside the object mask are made black, as in~\cite{Ausserlechner2023ZS6DZ6,nguyen2023cnos}. As shown on rows 8 vs. 9 in Tab.~\ref{tab:ablations}, this modification improves the performance of the\;\texttt{cls}-based approach but is still far from the performance of our bag-of-words approach (row 1).
When tried with the\;\texttt{cls}\;token from layer 23, the alternative retrieval approach achieved even lower accuracy.

We further analyze the quality of templates retrieved by the bag-of-words approach by evaluating the object pose associated with the top retrieved template. The pose from the template is adjusted such that the 2D bounding circle of the object is aligned with the 2D bounding circle of the object segmentation mask in the query image. This approach is surprisingly effective, reaching accuracy close to the coarse poses from MegaPose while being 15 times faster (row 10 in Tab.~\ref{tab:ablations} vs row 4 in Tab.~\ref{tab:main_results}).

Another possible retrieval approach is to directly match patches of each query-template pair, and calculate the pair similarity from the matching quality, \eg, as done by Goodwin \etal \cite{goodwin2022}. This approach needs a kNN search for each pair and is noticeably less efficient than our approach (0.64s vs 0.0008s per query; measured with\;\texttt{faiss}\cite{johnson2019billion}).

\myparagraph{Ground-truth segmentation masks.}
A common source of failure cases are erroneous segmentation masks. When ground-truth masks are used instead of masks predicted by CNOS~\cite{nguyen2023cnos}, FoundPose achieves a large +6--19 AR improvement (rows 1 vs 11 in Tab.~\ref{tab:ablations}).

\myparagraph{Effect of other parameters.}
We also evaluated FoundPose with different numbers of templates (400, 800, 1600), PCA components (128, 256, 512, 1024), visual words (1024, 2048, 4096), and RANSAC iterations (200, 400, 800). These settings influence memory requirements but lead to only marginal differences in speed and accuracy.

\myparagraph{Memory requirements.}
With around 300 valid patches per template, patch descriptors projected to a 256$d$ PCA space, and 4B per scalar, the patch descriptors take 300kB per template. With 2048 visual words, the bag-of-words descriptor takes an additional 8kB per template. In total, an object representation built from 800 templates takes 234MB. Compared to previous template-based approaches, this is \emph{25X less memory} than required by OSOP \cite{shugurov2022osop} and Nguyen \etal \cite{nguyen2022template}, which both need 5.6GB per object (16384$d$ vector for each of 92232 templates). With a sub 1\;AR drop, the memory required by FoundPose can be reduced to 59MB with 400 templates, 128$d$ PCA space, and 1024 visual words.
The template-based method of Sundermeyer \etal\cite{sundermeyer2020} requires 45MB per object, but its accuracy is -13.1 AR lower compared to FoundPose (see \cite{sundermeyer2020} for details). The concurrent works GigaPose~\cite{nguyen2024gigaPose} and ZS6D\cite{Ausserlechner2023ZS6DZ6} have similar per-template memory requirements as FoundPose, use 162 and 300 templates, respectively but achieve lower accuracy (Tab.~\ref{tab:main_results}).

%% file: tex/quals.tex
\begin{figure}[t!]
  \begin{center}
   \includegraphics[width=.31\linewidth]{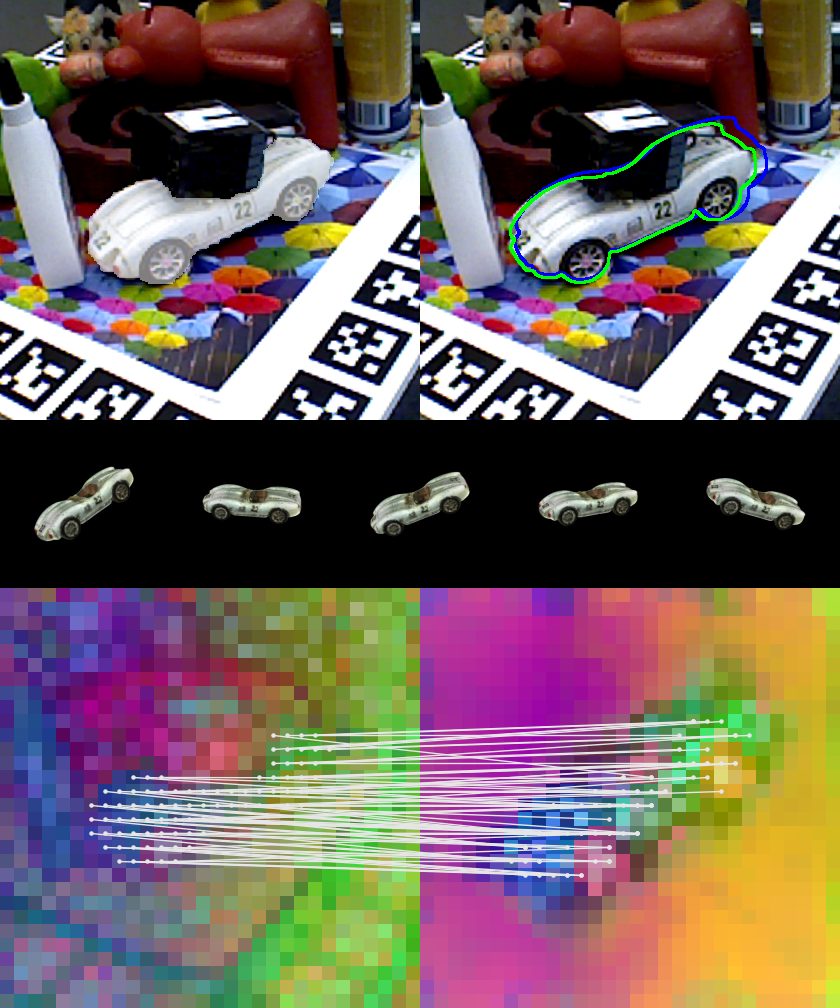} \hspace{0.1ex}
   \includegraphics[width=.31\linewidth]{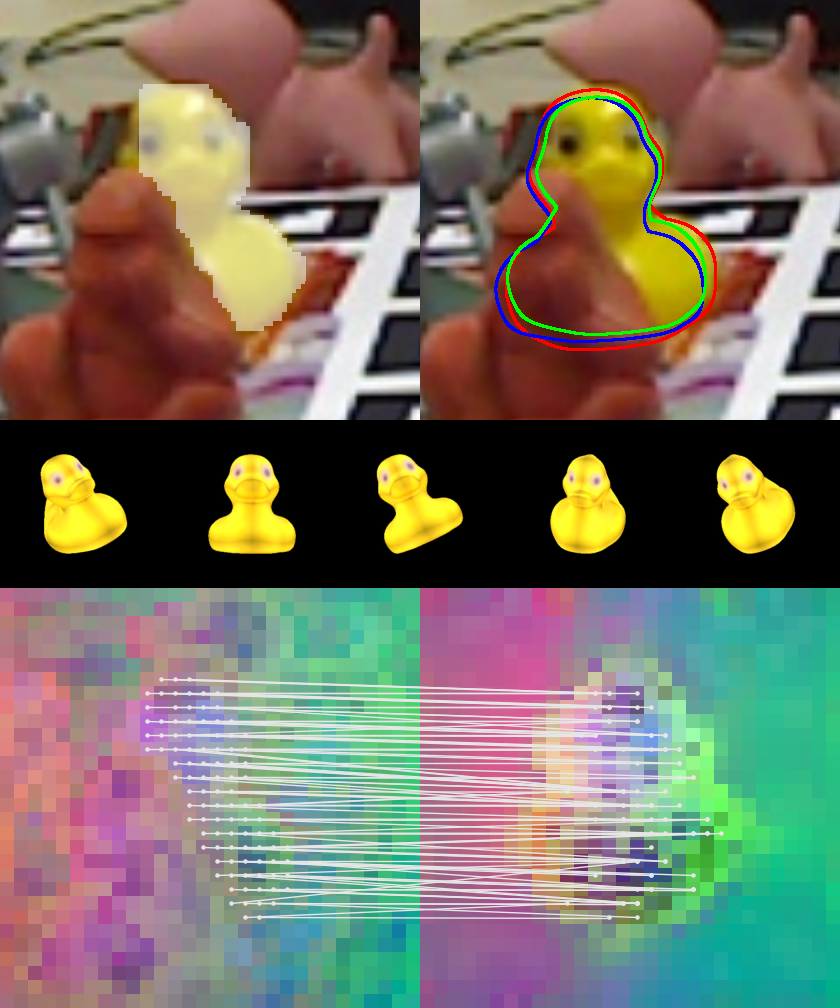} \hspace{0.1ex}
   \includegraphics[width=.31\linewidth]{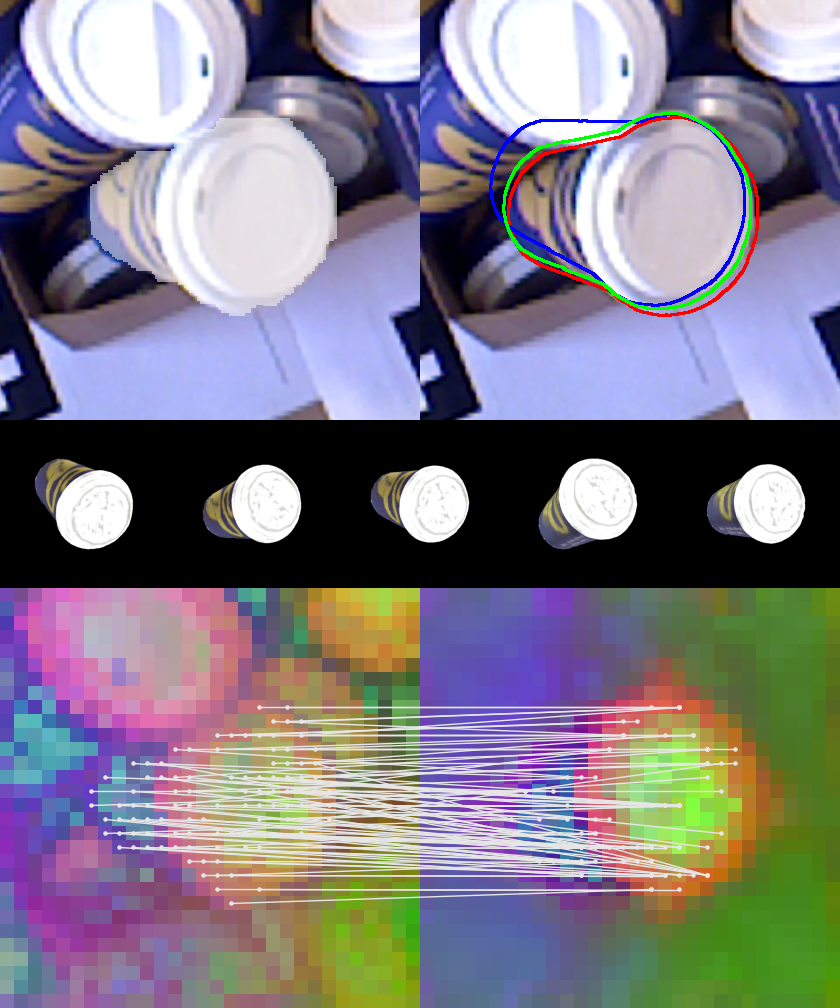} \hspace{0.1ex} \\ \vspace{1.38ex}
   \includegraphics[width=.31\linewidth]{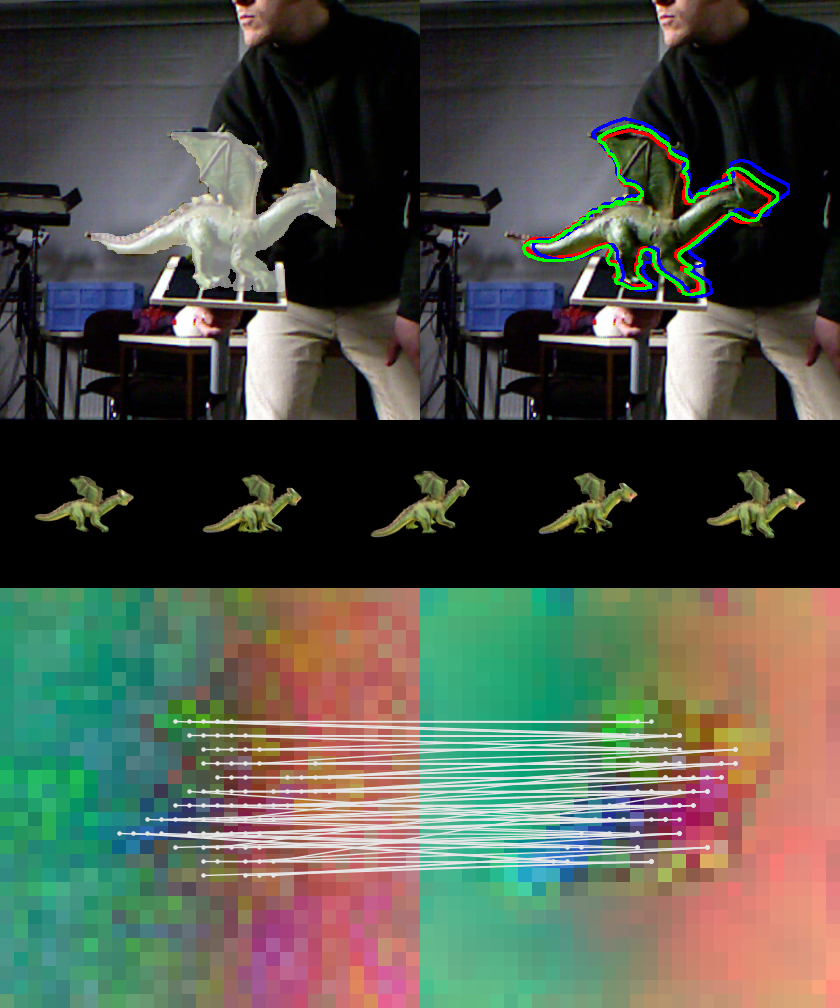} \hspace{0.1ex}
   \includegraphics[width=.31\linewidth]{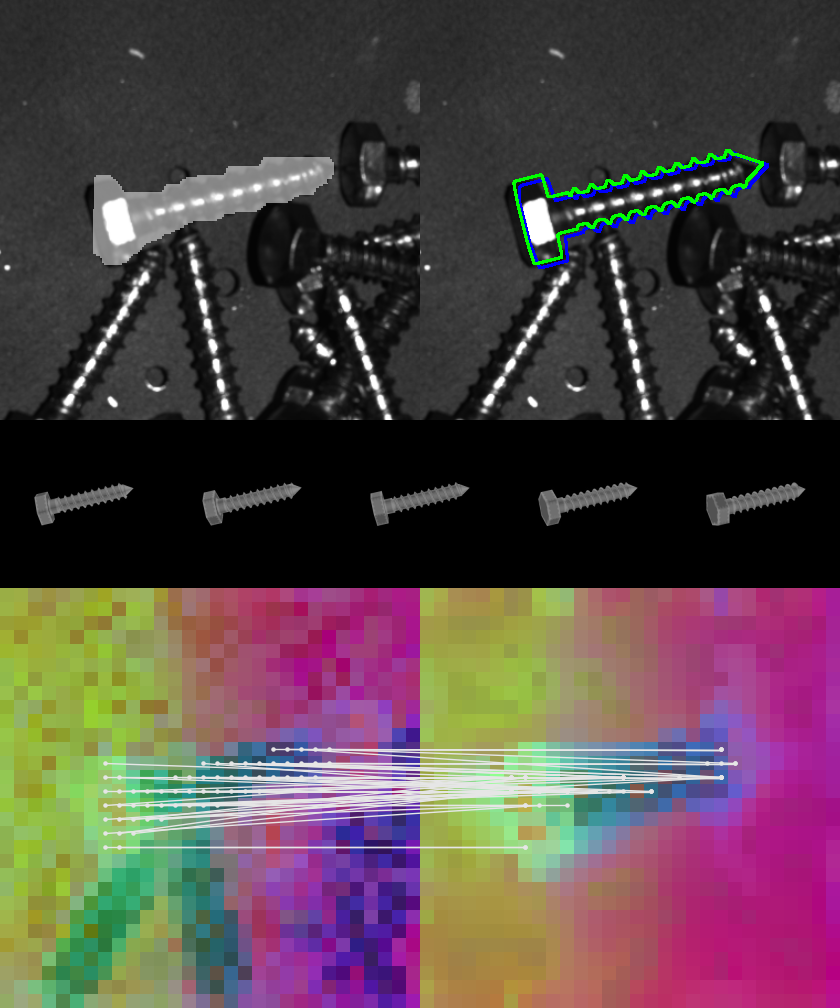} \hspace{0.1ex}
   \includegraphics[width=.31\linewidth]{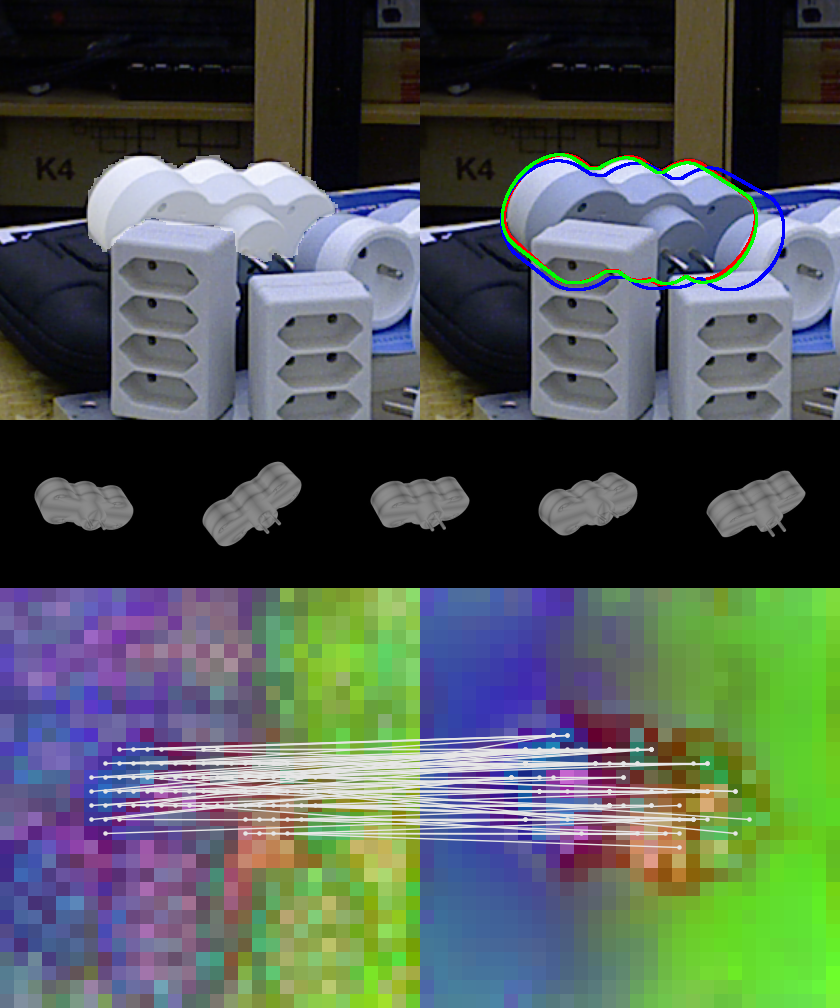} \hspace{0.1ex}
   \\
   \caption{\textbf{Example FoundPose results} on HB, LM-O, IC-BIN, TUD-L, ITODD and T-LESS datasets, showing that our method can handle a broad range or objects, including textured, texture-less and symmetric ones. Each example shows the query image crop with the CNOS mask in white (top left), retrieved templates (middle row), matched patch descriptors of the crop and the template that led to the top-quality pose estimate
   (bottom row), and the contour of the ground-truth pose in red, the coarse pose in blue, and the refined pose in green (top right).
   }
   \label{fig:quals}
   \vspace{-1.3em}
   \end{center}
\end{figure}

%% file: tex/mainresults.tex
\setlength{\tabcolsep}{2.5pt}
\begin{table}[t!]
    \begin{center}
    \scriptsize
    \begin{tabularx}{1.0\linewidth}{r l l Y Y Y Y Y Y Y Y Y Y}
    \toprule
       \# \vspace{0.4ex} &
       Method &
       Pose refinement &
       \rotatebox[origin=lB]{90}{No train.} &
       \rotatebox[origin=lB]{90}{LM-O} &
       \rotatebox[origin=lB]{90}{T-LESS} &
       \rotatebox[origin=lB]{90}{TUD-L} &
       \rotatebox[origin=lB]{90}{IC-BIN} &
       \rotatebox[origin=lB]{90}{ITODD} &
       \rotatebox[origin=lB]{90}{HB} &
       \rotatebox[origin=lB]{90}{YCB-V} &
       \rotatebox[origin=lB]{90}{Average} &
       \rotatebox[origin=lB]{90}{Time}\\
    \toprule
    \multicolumn{13}{l}{\textit{Coarse pose estimation:}} \\
    \midrule

    \Rowcolor{lightergray} \phantom{0}1 & FoundPose & -- & \cmark & \textbf{39.6}	& \textbf{33.8}	& \textbf{46.7}	& \textbf{23.9}	& \textbf{20.4}	& \textbf{50.8}	& \textbf{45.2}	& \textbf{37.2}	& \phantom{0}\underline{1.7} \\

    2 & GigaPose \cite{nguyen2024gigaPose} & -- & \xmark & 29.9 & \underline{27.3} & \underline{30.2} & \underline{23.1} & \underline{18.8} & 34.8 & 29.0 & \underline{27.6} & \phantom{0}\textbf{0.9}  \\ 
    
    3 & GenFlow \cite{genflow} & -- & \xmark & 25.0 & {21.5} & {30.0} & {16.8} & {15.4} & 28.3 & 27.7 & {23.5} & \phantom{0}3.8 \\
   
    4 & MegaPose \cite{labbe2022megapose} & -- & \xmark & 22.9 & 17.7 & 25.8 & 15.2 & 10.8 & 25.1 & 28.1 & 20.8 & 15.5\\
    
    5 & OSOP \cite{shugurov2022osop} & -- & \xmark & \underline{31.2} & -- & -- & -- & --  & \underline{49.2}  & \underline{33.2} & -- &  -- \\

    6 & ZS6D \cite{Ausserlechner2023ZS6DZ6} & --  & \cmark & 29.8 & 21.0 & -- & -- & -- & -- & {32.4} & -- &  -- \\
    
    \midrule
    \multicolumn{13}{l}{\textit{With pose refinement (a single hypothesis):}} \\
    \midrule

    \Rowcolor{lightergray} \phantom{0}7 & FoundPose & Featuremetric & \cmark & 39.5 & 39.6	& 56.7	& 28.3	& 26.2	& 58.5	& 49.7	& 42.6	& \phantom{0}\underline{2.6} \\

    \Rowcolor{lightergray} \phantom{0}8 & FoundPose & MegaPose & \xmark & {55.4}	& \underline{51.0}	& \underline{63.3}	& {43.0}	& {34.6}	& \underline{69.5}	& \textbf{66.1}	& \underline{54.7}	& \phantom{0}{4.4}\\

     \Rowcolor{lightergray} \phantom{0}9 & FoundPose & Feat.\,+\,MegaPose
     & \xmark & \textbf{55.7}	& \underline{51.0}	& \underline{63.3}	& \underline{43.3}	& \underline{35.7}	& \textbf{69.7}	& \textbf{66.1}	& \textbf{55.0}	& \phantom{0}6.4 \\

    10 & Gigapose \cite{nguyen2024gigaPose} & MegaPose & \xmark & \underline{55.6} & \textbf{54.6} & 57.8 & \textbf{44.3} & \textbf{37.8} & 69.3 & \underline{63.4} & \underline{54.7} & \phantom{0}\textbf{2.4} \\
    
    11 & MegaPose \cite{labbe2022megapose} & MegaPose & \xmark & 49.9 & {47.7} & \textbf{65.3} & 36.7 & 31.5 & 65.4 & {60.1} & 50.9 & 31.7 \\
    
    \midrule
    \multicolumn{13}{l}{\textit{With pose refinement (5 hypotheses):}} \\
    \midrule

    \Rowcolor{lightergray} 12 & FoundPose & Featuremetric & \cmark & 42.0	& 43.6	& 60.2	& 30.5	& 27.3	& 53.7	& 51.3	& 44.1	& \phantom{0}\underline{7.4}\\

     \Rowcolor{lightergray} 13 &  FoundPose & MegaPose & \xmark & 58.6	& \underline{54.9}	& 65.7	& 44.4	& 36.1	& 70.3	& \underline{67.3}	& 56.8	& {11.2} \\ 

    \Rowcolor{lightergray} 14 &  FoundPose & Feat.\,+\,MegaPose & \xmark & \textbf{61.0}	& \textbf{57.0} &	\textbf{69.4}	& \textbf{47.9}	& \textbf{40.7}	& \underline{72.3}	& \textbf{69.0}	& \textbf{59.6}	& 20.5\\

    15 & GigaPose \cite{nguyen2024gigaPose} & MegaPose & \xmark & \underline{59.9} & \textbf{57.0} & 64.5 & \underline{46.7} & \underline{39.7} & 72.2 & 66.3 & \underline{57.9} & \phantom{0}\textbf{7.3}\\

    16 & GenFlow \cite{genflow} & GenFlow & \xmark & 56.3	& 52.3	& \underline{68.4}	& 45.3	& 39.5	& \textbf{73.9}	& 63.3	& 57.1 & {20.9} \\

    17 & MegaPose \cite{labbe2022megapose} & MegaPose & \xmark & 56.0	& 50.7	& \underline{68.4}	& 41.4	& 33.8	& 70.4	& 62.1	& 54.7 & 47.4 \\

    \bottomrule
    \end{tabularx}
    \end{center}
    \caption{\textbf{Performance on the seven core BOP datasets~\cite{sundermeyer2022bop}.} The table shows Average Recall (AR) scores per dataset, the average AR score, and the time to estimate poses of all objects in an image averaged over the datasets (in seconds). Methods for coarse pose estimation (without applying any refinement) are in the upper part, methods applying a refinement stage on a single pose hypothesis are in the middle, and methods refining multiple pose hypotheses and reporting the top refined pose (with the highest refinement score) are at the bottom.
    Methods that do not undergo any task-specific training are marked with \cmark.
    }
    \label{tab:main_results}
\end{table}

%% file: tex/ablations.tex
\begin{table}[t!]
    \setlength{\tabcolsep}{2.5pt}
    \scriptsize %
    \begin{center}
    \begin{tabularx}{1.0\linewidth}{r l Y Y Y Y Y Y Y Y Y}
    \toprule
    \# \vspace{0.4ex} &
    Method &
    \rotatebox[origin=lB]{90}{LM-O} &
    \rotatebox[origin=lB]{90}{T-LESS} &
    \rotatebox[origin=lB]{90}{TUD-L} &
    \rotatebox[origin=lB]{90}{IC-BIN} &
    \rotatebox[origin=lB]{90}{ITODD} &
    \rotatebox[origin=lB]{90}{HB} &
    \rotatebox[origin=lB]{90}{YCB-V} &
    \rotatebox[origin=lB]{90}{Average} &
    \rotatebox[origin=lB]{90}{Time} \\
    \toprule
    
    & \multicolumn{8}{l}{\textit{Backbones for extracting patch descriptors:}} \\
    \midrule

    \Rowcolor{lightergray} \phantom{0}1 & DINOv2 ViT-L -- layer 18 & 39.6	& 33.8	& 46.7	& 23.9	& 20.4	& 50.8	& 45.2	& 37.2	& 1.7\\
    
    2 & DINOv2 ViT-L -- layer 23 &  23.2	& 22.8	& 31.2	& 10.3	& \phantom{0}9.7	& 33.0	& 34.0	& 23.5	& 1.5\\

    3 & DINOv2 ViT-S -- layer 9 & 34.0	& 31.6	& 42.7	&21.7	& 16.8	& 46.8	& 44.7	& 34.0	& 1.3\\

    4 & DINOv2 ViT-S -- layer 11 & 22.8	& 24.2	& 29.8	& 11.9	& 10.5	& 30.4	& 36.4	& 23.7	& 1.3\\

    5 & SAM ViT-L \cite{kirillov2023segany} -- layer 23 & \phantom{0}2.2	& 12.8	& \phantom{0}9.2	& \phantom{0}7.5	& \phantom{0}6.0 & 10.6	& 26.9	& 10.7	& 3.4 \\

    6 & DenseSIFT  -- step size 7px  & \phantom{0}3.2	& \phantom{0}2.6 & \phantom{0}6.5	& 10.5	& \phantom{0}2.9	& \phantom{0}5.6	& 22.2	& \phantom{0}7.6	& 1.4 \\

    7 & S2DNet\cite{germain2020s2dnet} & \phantom{0}0.8	& \phantom{0}1.2	& \phantom{0}0.8	& \phantom{0}1.4	& \phantom{0}1.2	& \phantom{0}1.2	& \phantom{0}1.3	& \phantom{0}1.1	& 1.8 \\

    \midrule
    & \multicolumn{8}{l}{\textit{Template retrieval by matching\;\texttt{cls}\;token from layer 18 of DINOv2 ViT-L:}} \\
    \midrule

    8 & Retrieval by\;\texttt{cls}\;token & 19.9	& 17.8	& 24.6	& 10.3	& 13.6	& 17.7	& 23.6	& 18.2	& 1.6 \\

    9 & Retrieval by\;\texttt{cls}\;token with black bg. & 25.5	& 26.2	& 30.3	& 16.7	& 13.6	& 29.3	& 34.4	& 25.1	& 1.6 \\

    \midrule
    & \multicolumn{8}{l}{\textit{Other ablations:}} \\
    \midrule

    10 & Pose given by the top matched template & 20.3	& 18.5	& 23.0 & 12.8	& 12.4	& 19.6	& 17.6	& 17.7	& 1.0 \\
    11 & Ground-truth instead of CNOS masks & 45.6	& 53.1	& 57.1	& 30.6 & --	& --	& 50.9 & -- & -- \\
    \bottomrule
    \end{tabularx}
    \end{center}
    \caption{\textbf{Ablation experiments.} We analyze different backbones for extracting patch descriptors (rows 1--7), compare our bag-of-words template retrieval with alternative approaches based on matching the\;\texttt{cls}\;token (rows 1 vs 8 and 9), evaluate our retrieval approach by considering the pose associated with the top template as the final pose estimate (row 10), and show the accuracy of coarse poses from FoundPose when ground-truth masks are used instead of the CNOS masks\cite{nguyen2023cnos} (row 11; ground-truth annotations for ITODD and HB are not publicly available).
    }
    \label{tab:ablations}
\end{table}

%% file: sections/5_conclusion.tex
\section{Conclusion}
\label{sec:concl}

We have proposed an RGB method for model-based pose estimation of unseen objects, which significantly outperforms existing methods on the standard BOP benchmark. We believe that achieving this without any object- nor task-specific training, just with a frozen vision foundation model, is an important and non-obvious outcome. Furthermore, we have shown that the method can be seamlessly combined with an existing render-and-compare refinement approach to achieve RGB-only state-of-the-art results. Our strong results are encouraging to revisit efficient classical computer vision which is often overlooked in the modern literature.

%% file: main.bbl
\begin{thebibliography}{10}
\providecommand{\url}[1]{\texttt{#1}}
\providecommand{\urlprefix}{URL }
\providecommand{\doi}[1]{https://doi.org/#1}

\bibitem{alexa2022super}
Alexa, M.: {Super-Fibonacci Spirals}: Fast, low-discrepancy sampling of {SO(3)}. CVPR  (2022)

\bibitem{amir2021deep}
Amir, S., Gandelsman, Y., Bagon, S., Dekel, T.: Deep {ViT} features as dense visual descriptors. ECCVW  (2022)

\bibitem{Ausserlechner2023ZS6DZ6}
Ausserlechner, P., Haberger, D., Thalhammer, S., Weibel, J.B., Vincze, M.: {ZS6D}: Zero-shot {6D} object pose estimation using vision transformers. arXiv preprint arXiv:2309.11986  (2023)

\bibitem{baker2004lucas}
Baker, S., Matthews, I.: {Lucas-Kanade} 20 years on: A unifying framework. IJCV  (2004)

\bibitem{Balntas2017PoseGR}
Balntas, V., Doumanoglou, A., Sahin, C., Sock, J., Kouskouridas, R., Kim, T.K.: Pose guided {RGBD} feature learning for {3D} object pose estimation. ICCV  (2017)

\bibitem{barronloss}
Barron, J.T.: A general and adaptive robust loss function. CVPR  (2019)

\bibitem{bommasani2021opportunities}
Bommasani, R., Hudson, D.A., Adeli, E., Altman, R., Arora, S., von Arx, S., Bernstein, M.S., Bohg, J., Bosselut, A., Brunskill, E., et~al.: On the opportunities and risks of foundation models. arXiv preprint arXiv:2108.07258  (2021)

\bibitem{brachmann2014learning}
Brachmann, E., Krull, A., Michel, F., Gumhold, S., Shotton, J., Rother, C.: Learning {6D} object pose estimation using {3D} object coordinates. ECCV  (2014)

\bibitem{brown2020language}
Brown, T., Mann, B., Ryder, N., Subbiah, M., Kaplan, J.D., Dhariwal, P., Neelakantan, A., Shyam, P., Sastry, G., Askell, A., et~al.: Language models are few-shot learners. NeurIPS  (2020)

\bibitem{caraffa2023object}
Caraffa, A., Boscaini, D., Hamza, A., Poiesi, F.: Object {6D} pose estimation meets zero-shot learning (2023)

\bibitem{caron2021emerging}
Caron, M., Touvron, H., Misra, I., J\'egou, H., Mairal, J., Bojanowski, P., Joulin, A.: Emerging properties in self-supervised vision transformers. ICCV  (2021)

\bibitem{chen2023zeropose}
Chen, J., Sun, M., Bao, T., Zhao, R., Wu, L., He, Z.: {{ZeroPose}}: {CAD}-model-based zero-shot pose estimation. arXiv preprint arXiv:2305.17934  (2023)

\bibitem{cherti2023reproducible}
Cherti, M., Beaumont, R., Wightman, R., Wortsman, M., Ilharco, G., Gordon, C., Schuhmann, C., Schmidt, L., Jitsev, J.: Reproducible scaling laws for contrastive language-image learning. CVPR  (2023)

\bibitem{collet2011moped}
Collet, A., Martinez, M., Srinivasa, S.S.: The {{MOPED}} framework: Object recognition and pose estimation for manipulation. IJRR  (2011)

\bibitem{darcet2023vision}
Darcet, T., Oquab, M., Mairal, J., Bojanowski, P.: Vision transformers need registers. arXiv preprint arXiv:2309.16588  (2023)

\bibitem{denninger2020blenderproc}
Denninger, M., Sundermeyer, M., Winkelbauer, D., Olefir, D., Hodan, T., Zidan, Y., Elbadrawy, M., Knauer, M., Katam, H., Lodhi, A.: {BlenderProc:} reducing the reality gap with photorealistic rendering. Robotics: Science and Systems (RSS) Workshops  (2020)

\bibitem{devlin2019bert}
Devlin, J., Chang, M.W., Lee, K., Toutanova, K.: {{BERT}}: Pre-training of deep bidirectional transformers for language understanding. {{ACL}}  (2019)

\bibitem{dosovitskiy2021image}
Dosovitskiy, A., Beyer, L., Kolesnikov, A., Weissenborn, D., Zhai, X., Unterthiner, T., Dehghani, M., Minderer, M., Heigold, G., Gelly, S., Uszkoreit, J., Houlsby, N.: An image is worth 16x16 words: Transformers for image recognition at scale. {{ICLR}}  (2021)

\bibitem{doumanoglou2016recovering}
Doumanoglou, A., Kouskouridas, R., Malassiotis, S., Kim, T.K.: Recovering {6D} object pose and predicting next-best-view in the crowd. CVPR  (2016)

\bibitem{drost2017introducing}
Drost, B., Ulrich, M., Bergmann, P., Hartinger, P., Steger, C.: Introducing {MVTec} {ITODD} -- {A} dataset for {3D} object recognition in industry. ICCVW  (2017)

\bibitem{drost2010model}
Drost, B., Ulrich, M., Navab, N., Ilic, S.: Model globally, match locally: {E}fficient and robust {3D} object recognition. CVPR  (2010)

\bibitem{fischler1981random}
Fischler, M.A., Bolles, R.C.: Random sample consensus: {A} paradigm for model fitting with applications to image analysis and automated cartography. Communications of the ACM  (1981)

\bibitem{germain2020s2dnet}
Germain, H., Bourmaud, G., Lepetit, V.: {S2DNet}: Learning accurate correspondences for sparse-to-dense feature matching. ECCV  (2020)

\bibitem{goodwin2022}
Goodwin, W., Vaze, S., Havoutis, I., Posner, I.: Zero-shot category-level object pose estimation. ECCV  (2022)

\bibitem{he2017mask}
He, K., Gkioxari, G., Doll{\'a}r, P., Girshick, R.: Mask {R-CNN}. ICCV  (2017)

\bibitem{he2022oneposeplusplus}
He, X., Sun, J., Wang, Y., Huang, D., Bao, H., Zhou, X.: {OnePose++}: Keypoint-free one-shot object pose estimation without {CAD} models. NeurIPS  (2022)

\bibitem{he2022fs6d}
He, Y., Wang, Y., Fan, H., Sun, J., Chen, Q.: {{FS6D}}: Few-shot {6D} pose estimation of novel objects. {{CVPR}}  (2022)

\bibitem{hinterstoisser2012accv}
Hinterstoisser, S., Lepetit, V., Ilic, S., Holzer, S., Bradski, G., Konolige, K., Navab, N.: Model based training, detection and pose estimation of texture-less 3{D} objects in heavily cluttered scenes. ACCV  (2012)

\bibitem{hinterstoisser2010dominant}
Hinterstoisser, S., Lepetit, V., Ilic, S., Fua, P., Navab, N.: Dominant orientation templates for real-time detection of texture-less objects. {{CVPR}}  (2010)

\bibitem{hodan2020epos}
Hodan, T., Barath, D., Matas, J.: {EPOS}: Estimating {6D} pose of objects with symmetries. {{CVPR}}  (2020)

\bibitem{hodan2017tless}
Hodan, T., Haluza, P., Obdrzalek, S., Matas, J., Lourakis, M., Zabulis, X.: {T-LESS}: {A}n {RGB-D} dataset for {6D} pose estimation of texture-less objects. WACV  (2017)

\bibitem{hodan2018bop}
Hodan, T., Michel, F., Brachmann, E., Kehl, W., Glent~Buch, A., Kraft, D., Drost, B., Vidal, J., Ihrke, S., Zabulis, X., Sahin, C., Manhardt, F., Tombari, F., Kim, T.K., Matas, J., Rother, C.: {BOP}: {B}enchmark for {6D} object pose estimation. ECCV  (2018)

\bibitem{hodan2020bop}
Hodan, T., Sundermeyer, M., Drost, B., Labb\'{e}, Y., Brachmann, E., Michel, F., Rother, C., Matas, J.: {BOP} challenge 2020 on {6D} object localization. ECCVW  (2020)

\bibitem{hodan2023bop}
Hodan, T., Sundermeyer, M., Labbe, Y., Wang, G., Brachmann, E., Drost, B., Rother, C., Matas, J.: {BOP} challenge 2023 on detection, segmentation and pose estimation of unseen rigid objects. To be published.  (2023), \url{https://bop.felk.cvut.cz/leaderboards/pose-estimation-unseen-bop23/core-datasets/}, the results are available at bop.felk.cvut.cz.

\bibitem{hodan2019photorealistic}
Hoda{\v{n}}, T., Vineet, V., Gal, R., Shalev, E., Hanzelka, J., Connell, T., Urbina, P., Sinha, S., Guenter, B.: Photorealistic image synthesis for object instance detection. IEEE International Conference on Image Processing (ICIP)  (2019)

\bibitem{hodan2015detection}
Hoda{\v{n}}, T., Zabulis, X., Lourakis, M., Obdr{\v{z}}{\'a}lek, {\v{S}}., Matas, J.: Detection and fine 3d pose estimation of texture-less objects in {RGB-D} images. IEEE International Conference on Intelligent Robots and Systems (IROS)  (2015)

\bibitem{jia2021scaling}
Jia, C., Yang, Y., Xia, Y., Chen, Y.T., Parekh, Z., Pham, H., Le, Q.V., Sung, Y., Li, Z., Duerig, T.: Scaling up visual and vision-language representation learning with noisy text supervision. {{ICML}}  (2021)

\bibitem{johnson2019billion}
Johnson, J., Douze, M., J{\'e}gou, H.: Billion-scale similarity search with {GPUs}. IEEE Transactions on Big Data  (2019)

\bibitem{kaskman2019homebreweddb}
Kaskman, R., Zakharov, S., Shugurov, I., Ilic, S.: {HomebrewedDB}: {RGB-D} dataset for {6D} pose estimation of {3D} objects. ICCVW  (2019)

\bibitem{kirillov2023segany}
Kirillov, A., Mintun, E., Ravi, N., Mao, H., Rolland, C., Gustafson, L., Xiao, T., Whitehead, S., Berg, A.C., Lo, W.Y., Doll{\'a}r, P., Girshick, R.: Segment anything. ICCV  (2023)

\bibitem{labbe2022megapose}
Labb\'e, Y., Manuelli, L., Mousavian, A., Tyree, S., Birchfield, S., Tremblay, J., Carpentier, J., Aubry, M., Fox, D., Sivic, J.: {{MegaPose}}: {6D} pose estimation of novel objects via render \& compare. CoRL  (2022)

\bibitem{labbe2020cosypose}
Labbé, Y., Carpentier, J., Aubry, M., Sivic, J.: {{CosyPose}}: Consistent multi-view multi-object {6D} pose estimation. {{ECCV}}  (2020)

\bibitem{lepetit2009epnp}
Lepetit, V., Moreno-Noguer, F., Fua, P.: {EPnP:} {A}n accurate {O(n)} solution to the {PnP} problem. IJCV  (2009)

\bibitem{levenberg1944}
Levenberg, K.: A method for the solution of certain non-linear problems in least squares. Quarterly of Applied Mathematics  (1944)

\bibitem{li2020deepim}
Li, Y., Wang, G., Ji, X., Xiang, Y., Fox, D.: {{DeepIM}}: Deep iterative matching for {6D} pose estimation. IJCV  (2020)

\bibitem{lin2023sam}
Lin, J., Liu, L., Lu, D., Jia, K.: {SAM-6D}: Segment anything model meets zero-shot 6d object pose estimation. arXiv preprint arXiv:2311.15707  (2023)

\bibitem{liu2023matcher}
Liu, Y., Zhu, M., Li, H., Chen, H., Wang, X., Shen, C.: Matcher: Segment anything with one shot using all-purpose feature matching. arXiv preprint arXiv:2305.13310  (2023)

\bibitem{liu2022gen6d}
Liu, Y., Wen, Y., Peng, S., Lin, C., Long, X., Komura, T., Wang, W.: {Gen6D}: Generalizable model-free {6-DoF} object pose estimation from {RGB} images. ECCV  (2022)

\bibitem{lowe1999object}
Lowe, D.G.: Object recognition from local scale-invariant features. {{ICCV}}  (1999)

\bibitem{lowe2004distinctive}
Lowe, D.G.: Distinctive image features from scale-invariant keypoints. IJCV  (2004)

\bibitem{manhardt2018deep}
Manhardt, F., Kehl, W., Navab, N., Tombari, F.: Deep {{Model-Based 6D Pose Refinement}} in {{RGB}}. {{ECCV}}  (2018)

\bibitem{marquardt1963}
Marquardt, D.W.: An algorithm for least-squares estimation of nonlinear parameters. Journal of the Society for Industrial and Applied Mathematics  (1963)

\bibitem{melaskyriazi2022deep}
Melas-Kyriazi, L., Rupprecht, C., Laina, I., Vedaldi, A.: {Deep Spectral Methods}: A surprisingly strong baseline for unsupervised semantic segmentation and localization. CVPR  (2022)

\bibitem{genflow}
Moon, S., Son, H.: {{GenFlow}}, a submission to the {{BOP Challenge}} 2023 (bop.felk.cvut.cz). Unpublished  (2023)

\bibitem{murase1995visual}
Murase, H., Nayar, S.K.: Visual learning and recognition of {3-D} objects from appearance. IJCV  (1995)

\bibitem{3dczsl2022}
Naeem, M.F., {\"O}rnek, E.P., Xian, Y., Van~Gool, L., Tombari, F.: {3D} compositional zero-shot learning with decompositional consensus. In: ECCV (2022)

\bibitem{newcombe2011kinectfusion}
Newcombe, R.A., Izadi, S., Hilliges, O., Molyneaux, D., Kim, D., Davison, A.J., Kohi, P., Shotton, J., Hodges, S., Fitzgibbon, A.: {K}inect{F}usion: {R}eal-time dense surface mapping and tracking. ISMAR  (2011)

\bibitem{nguyen2023cnos}
Nguyen, V.N., Groueix, T., Ponimatkin, G., Lepetit, V., Hodan, T.: {{CNOS}}: A strong baseline for {CAD}-based novel object segmentation. {{ICCVW}}  (2023)

\bibitem{nguyen2024gigaPose}
Nguyen, V.N., Groueix, T., Salzmann, M., Lepetit, V.: {GigaPose}: Fast and robust novel object pose estimation via one correspondence. In: CVPR (2024)

\bibitem{nguyen2022template}
Nguyen, V.N., Hu, Y., Xiao, Y., Salzmann, M., Lepetit, V.: Templates for {3D} object pose estimation revisited: Generalization to new objects and robustness to occlusions. CVPR  (2022)

\bibitem{okorn2021zephyr}
Okorn, B., Gu, Q., Hebert, M., Held, D.: {{ZePHyR}}: Zero-shot pose hypothesis rating. {{ICRA}}  (2021)

\bibitem{oquab2023dinov2}
Oquab, M., Darcet, T., Moutakanni, T., Vo, H., Szafraniec, M., Khalidov, V., Fernandez, P., Haziza, D., Massa, F., El-Nouby, A., et~al.: {DINOv2}: Learning robust visual features without supervision. arXiv preprint arXiv:2304.07193  (2023)

\bibitem{ornek23}
{\"O}rnek, E.P., Krishnan, A.K., Gayaka, S., Kuo, C.H., Sen, A., Navab, N., Tombari, F.: {SupeRGB-D}: Zero-shot instance segmentation in cluttered indoor environments. IEEE RA-L  (2023)

\bibitem{park2020latentfusion}
Park, K., Mousavian, A., Xiang, Y., Fox, D.: {{LatentFusion}}: End-to-end differentiable reconstruction and rendering for unseen object pose estimation. {{CVPR}}  (2020)

\bibitem{philbin2007object}
Philbin, J., Chum, O., Isard, M., Sivic, J., Zisserman, A.: Object retrieval with large vocabularies and fast spatial matching. CVPR  (2007)

\bibitem{philbin2008lost}
Philbin, J., Chum, O., Isard, M., Sivic, J., Zisserman, A.: Lost in quantization: Improving particular object retrieval in large scale image databases. CVPR  (2008)

\bibitem{pitteri20203d}
Pitteri, G., Bugeau, A., Ilic, S., Lepetit, V.: {3D} object detection and pose estimation of unseen objects in color images with local surface embeddings. {{ACCV}}  (2020)

\bibitem{pitteri2019cornet}
Pitteri, G., Ilic, S., Lepetit, V.: {{CorNet}}: Generic {3D} corners for {6D} pose estimation of new objects without retraining. {{ICCVW}}  (2019)

\bibitem{ponce2004toward}
Ponce, J., Lazebnik, S., Rothganger, F., Schmid, C.: Toward true {3D} object recognition. In: Reconnaissance de Formes et Intelligence Artificielle (2004)

\bibitem{radford2021learning}
Radford, A., Kim, J.W., Hallacy, C., Ramesh, A., Goh, G., Agarwal, S., Sastry, G., Askell, A., Mishkin, P., Clark, J., Krueger, G., Sutskever, I.: Learning transferable visual models from natural language supervision. {{ICML}}  (2021)

\bibitem{reizenstein2021common}
Reizenstein, J., Shapovalov, R., Henzler, P., Sbordone, L., Labatut, P., Novotny, D.: Common objects in {3D}: Large-scale learning and evaluation of real-life {3D} category reconstruction. ICCV  (2021)

\bibitem{roberts1963machineperception}
Roberts, L.G.: Machine perception of three-dimensional solids. Ph.D. thesis, Massachusetts Institute of Technology (1963)

\bibitem{sarlin21pixloc}
Sarlin, P.E., Unagar, A., Larsson, M., Germain, H., Toft, C., Larsson, V., Pollefeys, M., Lepetit, V., Hammarstrand, L., Kahl, F., Sattler, T.: Back to the feature: Learning robust camera localization from pixels to pose. CVPR  (2021)

\bibitem{shreiner2009opengl}
Shreiner, D.: {OpenGL} programming guide: the official guide to learning {OpenGL}, versions 3.0 and 3.1. Pearson Education (2009)

\bibitem{shugurov2022osop}
Shugurov, I., Li, F., Busam, B., Ilic, S.: {OSOP}: A multi-stage one shot object pose estimation framework. CVPR  (2022)

\bibitem{sivic2003video}
Sivic, Zisserman: {Video Google}: A text retrieval approach to object matching in videos. ICCV  (2003)

\bibitem{stumberg2020}
von Stumberg, L., Wenzel, P., Yang, N., Cremers, D.: {LM-Reloc}: {Levenberg-Marquardt} based direct visual relocalization. 3DV  (2020)

\bibitem{sun2021loftr}
Sun, J., Shen, Z., Wang, Y., Bao, H., Zhou, X.: {LoFTR}: Detector-free local feature matching with transformers. CVPR  (2021)

\bibitem{sun2022onepose}
Sun, J., Wang, Z., Zhang, S., He, X., Zhao, H., Zhang, G., Zhou, X.: {OnePose}: One-shot object pose estimation without {CAD} models. CVPR  (2022)

\bibitem{sundermeyer2020multipath}
Sundermeyer, M., Durner, M., Puang, E.Y., Marton, Z.C., Vaskevicius, N., Arras, K.O., Triebel, R.: Multi-path learning for object pose estimation across domains. {{CVPR}}  (2020)

\bibitem{sundermeyer2020}
Sundermeyer, M., Durner, M., Puang, E.Y., Marton, Z.C., Vaskevicius, N., Kai, O.A., Triebel, R.: Multi-path learning for object pose estimation across domains. CVPR  (2020)

\bibitem{sundermeyer2022bop}
Sundermeyer, M., Hodan, T., Labb{\'e}, Y., Wang, G., Brachmann, E., Drost, B., Rother, C., Matas, J.: {BOP} challenge 2022 on detection, segmentation and pose estimation of specific rigid objects. CVPRW  (2023)

\bibitem{sundermeyer2018implicit}
Sundermeyer, M., Marton, Z.C., Durner, M., Brucker, M., Triebel, R.: Implicit {3D} orientation learning for {6D} object detection from rgb images. {{ECCV}}  (2018)

\bibitem{thalhammer2023selfsupervised}
Thalhammer, S., Weibel, J.B., Vincze, M., Garcia-Rodriguez, J.: Self-supervised vision transformers for {3D} pose estimation of novel objects. arXiv preprint arXiv:2306.00129  (2023)

\bibitem{tombari2013performance}
Tombari, F., Salti, S., Di~Stefano, L.: Performance evaluation of {3D} keypoint detectors. IJCV  (2013)

\bibitem{torii2013visual}
Torii, A., Sivic, J., Pajdla, T., Okutomi, M.: Visual place recognition with repetitive structures. CVPR  (2013)

\bibitem{wang2021gdrnet}
Wang, G., Manhardt, F., Tombari, F., Ji, X.: {{GDR-Net}}: Geometry-guided direct regression network for monocular {6D} object pose estimation. {{CVPR}}  (2021)

\bibitem{wang2022selfsupervised}
Wang, Y., Shen, X., Hu, S., Yuan, Y., Crowley, J., Vaufreydaz, D.: Self-supervised transformers for unsupervised object discovery using normalized cut. {{CVPR}}  (2022)

\bibitem{wen2023foundationpose}
Wen, B., Yang, W., Kautz, J., Birchfield, S.: {FoundationPose}: Unified {6D} pose estimation and tracking of novel objects (2023)

\bibitem{wu2023multiview}
Wu, C.Y., Johnson, J., Malik, J., Feichtenhofer, C., Gkioxari, G.: Multiview compressive coding for {3D} reconstruction. CVPR  (2023)

\bibitem{xiang2018posecnn}
Xiang, Y., Schmidt, T., Narayanan, V., Fox, D.: {{PoseCNN}}: A convolutional neural network for {6D} object pose estimation in cluttered scenes. {{RSS}}  (2018)

\bibitem{Xiao2019PoseFromShape}
Xiao, Y., Qiu, X., Langlois, P., Aubry, M., Marlet, R.: Pose from shape: Deep pose estimation for arbitrary {3D} objects. BMVC  (2019)

\bibitem{zhang2023tale}
Zhang, J., Herrmann, C., Hur, J., Cabrera, L.P., Jampani, V., Sun, D., Yang, M.H.: A tale of two features: Stable diffusion complements {DINO} for zero-shot semantic correspondence. NeurIPS  (2023)

\end{thebibliography}
